\documentclass[12pt]{article}
\usepackage{amsmath,amsthm,amssymb,bm}
\usepackage{graphicx,psfrag,epsf}%,algpseudocode}
\usepackage{enumerate,multirow,array,color,natbib,booktabs,setspace}
\PassOptionsToPackage{hyphens}{url}
\usepackage[font=small,labelfont=bf]{caption}
\usepackage[total={6.45in,8.75in}]{geometry}
\usepackage{algorithm2e}
\usepackage{algorithmic}
\RequirePackage[colorlinks,citecolor=blue,urlcolor=blue]{hyperref}
\newcommand{\blind}{1}
\def\floor#1{\lfloor #1 \rfloor}
\newtheorem{thm}{Theorem}
\newtheorem{coro}{Corollary}

\newtheorem{lemma}{Lemma}

\newtheorem{example}{Example}
\newtheoremstyle{exampstyle}
{\topsep} % Space above
{\topsep} % Space below
{} % Body font
{} % Indent amount
{\bfseries} % Theorem head font
{.} % Punctuation after theorem head
{.5em} % Space after theorem head
{} % Theorem head spec (can be left empty, meaning `normal')

\let\proglang=\textsf

\newcommand{\Mean}{{\mbox{E}}}

\newcommand{\Cov}{{\mbox{cov}}}

\newcommand{\prob}{{\mbox{Pr}}}

\DeclareMathOperator*{\argmax}{arg\,max}
\DeclareMathOperator*{\argmin}{arg\,min}

\def\spacingset#1{\renewcommand{\baselinestretch}%
	{#1}\small\normalsize} \spacingset{1}

\newcommand{\change}[1]{{\leavevmode\color{black}{#1}}}

\usepackage{xr}
\begin{document}

%%%%%%%%%%%%%%%%%%%%%%%%%%%%%%%%%%%%%%%%%%%%%%%%%%%%%%%%%%%%%%%%%%%%%%%%%%%%%%

\if1\blind
{
\title{\Large{\textbf{Statistically Efficient Advantage Learning for Offline Reinforcement Learning in Infinite Horizons}}}
\author{
\large{Chengchun Shi, Shikai Luo, Yuan Le, Hongtu Zhu and  Rui Song} \\
\\
\normalsize{\textit{London School of Economics and Political Science}}\\
\normalsize{\textit{ByteDance}}\\
\normalsize{\textit{Shanghai University of Finance and Economics}}\\
\normalsize{\textit{University of North Carolina at Chapel Hill}}\\
\normalsize{and \textit{North Carolina State University}}
}
\date{}
\maketitle
} \fi

\if0\blind
{
\title{\Large{\textbf{Statistically Efficient Advantage Learning for Offline Reinforcement Learning in Infinite Horizons}}}
\author{
}
\date{}
\maketitle
} \fi

\bigskip
\begin{abstract}
We consider reinforcement learning (RL) methods  in offline domains without additional online data collection, such as mobile health applications. Most of  existing policy optimization algorithms in the computer science literature are developed in online settings where data are easy to collect or simulate. Their generalizations to mobile health applications with a pre-collected offline dataset remain unknown. The aim of this paper is to develop a novel advantage learning framework in order to efficiently use pre-collected data for policy optimization. %The key ingredient of our framework lies in the  construction of pseudo outcomes that are asymptotically unbiased to the optimal advantage function. This allows us to we cast RL into a supervised learning problem, which greatly improves the sample efficiency of the resulting algorithm. 
%The proposed method is motivated by a line of research on developing doubly-robust estimation methods for learning optimal individualized treatment regimes in the statistics literature. It takes any existing state-of-the-art Q-learning type algorithms as the input, and 
%justified by theoretical results, synthetic datasets, and a real dataset. 
The proposed method takes an optimal Q-estimator computed by any existing state-of-the-art RL algorithms as input, and outputs a new policy whose value is guaranteed to converge at a faster rate than the policy derived based on the initial Q-estimator. Extensive numerical experiments are conducted to back up our theoretical findings. A \proglang{Python} implementation of our proposed method is available at \url{https://github.com/leyuanheart/SEAL}.
\end{abstract}

\vfill

\newpage
 
\baselineskip=20pt

\section{Introduction}
\label{secintroduction}
Reinforcement learning \citep[RL, see][for an overview]{Sutton2018} is concerned with how intelligence agents learn and take actions in an unknown environment in order to maximize the cumulative reward that it receives. It has been arguably one of the most vibrant research frontiers in machine learning over the last few years. According to Google Scholar, over 40K scientific articles have been published in 2020 with the phrase ``reinforcement learning". 
Over 100 papers on RL were accepted for presentation at ICML 2021, a premier conference in the machine learning area, accounting for more than 10\% of the accepted papers in total. RL algorithms have been applied in a wide variety of real applications, including games \citep{silver2016mastering}, robotics \citep{kormushev2013reinforcement}, healthcare \citep{komorowski2018artificial}, bidding \citep{jin2018real}, ridesharing \citep{xu2018large} and automated driving \citep{de2019causal}, to name  a few.  

This paper is partly motivated by developing statistical learning methodologies in offline RL domains such as mobile health (mHealth). %and other related applications in healthcare. 
mHealth technologies have recently emerged due to the use of mobile phones, tablets computers or wearable devices. They play an important role in precision medicine as they offer a means to monitor a patient's health status and deliver interventions in real-time. They also collect rich longitudinal data for optimal treatment decision making. One motivating example being considered in this paper uses the OhioT1DM Dataset \citep{marling2018ohiot1dm}. %Public access to this dataset is provided. 
%This dataset is publicly available to researchers interested in improving the health and wellbeing of people with type 1 diabetes, an autoimmune disease wherein the pancreas produces insufficient levels of insulin. 
It contains 8 weeks of data for 6 patients with type 1 diabetes, an autoimmune disease wherein the pancreas produces insufficient levels of insulin. For those patients, their continuous glucose monitoring blood glucose levels, insulin doses being injected, self-reported times of meals and exercises are continually measured. Their outcomes have the potential to be improved by treatment policies tailored to the continually evolving health status of each patient \citep{luckett2019,shi2020value}. 

Despite the popularity of developing various RL algorithms in the computer science literature, statistics as a field, has only recently begun to engage with RL both in depth and in breadth. Most works in the statistics literature focused on developing data-driven methodologies for precision medicine with only a few treatment stages \citep[see e.g.,][]{Murphy2003,robins2004,chakraborty2010inference,qian2011,zhang2013robust,zhao2015,Wallace2015,song2015penalized,Alex2016,zhangyc2016,zhu2017greedy,shi2018maximin,wang2018quantile,qi2020multi,nie2020learning}. These methods require a large number of patients in the observed data to be consistent. They are not applicable to mHealth applications with only a few patients, which is the case in the OhioT1DM dataset. Nor are they applicable to many other sequential decision making problems in infinite horizons where the number of decision stages is allowed to diverge to infinity, such as games or robotics. Recently, a few algorithms have been proposed in the statistics literature for policy optimization in mHealth applications \citep{Ertefaie2018,luckett2019,hu2020personalized,liao2020batch,zhou2021estimating}. 

Among all existing methods in infinite horizons, Q-learning \citep{watkins1992q} is arguably one of the most popular model-free RL algorithms. It derives the optimal policy by learning an optimal Q-function,  without explicitly modelling the system dynamics. Variants of Q-learning include  gradient Q-learning \citep[][]{maei2010,Ertefaie2018}, fitted Q-iteration \citep[][]{riedmiller2005}, deep Q-network \citep[DQN,][]{mnih2015human}, double DQN \citep{van2016deep} %double Q-learning \citep[][]{hasselt2010double,van2016deep},  
and quantile DQN \citep{dabney2018distributional}, 
among others. %These Q-learning type algorithms derive the optimal policy by learning an optimal Q-function. 
All these Q-learning type algorithms are primarily motivated by the application of developing artificial intelligence in online video games, so their generalization to offline applications with a pre-collected dataset remains unknown. 

Different from online settings (e.g., video games) where data are easy to collect or simulate, the number of observations in many offline applications (e.g., healthcare) is limited. Take the OhioT1DM dataset as an example, only a few thousands observations are available \citep{shi2020does}. With such limited data, it is critical to develop RL algorithms that are {\it statistically efficient}. Instead of proposing a specific algorithm for policy optimization, our work undertakes the ambitious task of devising an ``efficiency enhancement" method that is generally applicable to any Q-learning type algorithms to improve their statistical efficiency. The input of our method is an optimal Q-estimator computed by existing state-of-the-art RL algorithms and the output is a new policy whose value converges at a faster rate than the policy derived based on the initial Q-estimator. 
%have made tremendous achievements in developing artificial intelligence in video games, they typically require tens to hundreds of millions of samples. In this paper, we aim to develop RL algorithms that efficiently use historical data for policy optimization in real-world settings, where data collection is expensive. 

The proposed method is motivated by a line of research on developing A-learning type algorithms\footnote{Similar algorithms are developed in the causal inference literature for heterogeneous treatment effects estimation \citep[see e.g.,][]{tian2014simple,nie2017quasi,kennedy2020optimal,li2021robust}.} to learn an optimal dynamic treatment regime (DTR) to implement precision medicine \citep[see e.g.,][]{Murphy2003,robins2004,lu2013variable}. These methods directly model the difference between two conditional mean functions (known as the contrast function). They are semi-parametrically efficient and outperform Q-learning \footnote{Q-learning here is different from those Q-learning type algorithms in RL, due to different data structures and model setups. It relies on a backward induction algorithm to identify the optimal DTR in finite horizon settings with only a few treatment stages. In contrast, Q-learning type algorithms in RL usually rely on a Markov assumption to derive the optimal policy in infinite horizons.}\citep[see e.g.,][]{chakraborty2010inference,qian2011} in cases where the Q-function is misspecified \citep{shi2018high}. In addition, A-learning has the so-called doubly robustness property, i.e., the estimated optimal DTR is consistent when either the model for the conditional mean function or the treatment assignment mechanism is correctly specified.

The contributions of our paper are summarized as follows. Methodologically, we propose a statistically efficient advantage learning procedure to estimate the optimal policy in offline infinite horizon settings. %Our contributions are multi-fold. Methodologically, %we propose a supervised learning framework for batch reinforcement learning. 
Our proposal integrates existing policy optimization and policy evaluation algorithms in RL. Specifically, %instead of using the Bellman equation to recursively update the Q-function, %to recursively update the Q-function, 
we start with applying existing Q-learning type algorithms to compute an initial estimator for the optimal Q-function. Based on these Q-estimators, we leverage ideas from the off-policy evaluation literature %on developing doubly-robust value estimates 
\citep[OPE, see e.g.][]{jiang2016,thomas2016data,liu2018,kallus2019efficiently,kallus2020double,shi2021deeply} to construct pseudo outcomes that are asymptotically unbiased to the optimal contrast function (see Section \ref{sec:optQcon} for the detailed definition). With these pseudo outcomes as the prediction target, we can directly apply existing state-of-the-art supervised learning algorithms to derive the optimal policy. The use of OPE effectively alleviates the bias of the estimated contrast function resulting from the potential model misspecification of the optimal Q-function, which in turn improves the statistical efficiency over Q-learning. In that sense, our proposal shares similar spirits with the A-learning type methods to learn DTRs in finite horizons.
%type algorithms. 
%estimate the optimal policy. This greatly improves the sample efficiency in learning.
%Then we can treat this vector as the state variable to estimate the optimal policy.

Theoretically, we show our estimated contrast function converges at a faster rate than the Q-function computed by existing state-of-the-art Q-learning type algorithms (Theorem \ref{thm:3}). This in turn implies that our estimated policy achieves a larger value function (Theorem \ref{thm:4}). All the error bounds derived in this paper  converge to zero when either 
%All our theoretical findings are derived under the bidirectional framework that requires either 
the number of trajectories $N$ or the number of decision stages per trajectory $T$ to approach infinity. This guarantees the consistency of our method when applied to a wide range of real-world  problems, ranging from the OhioT1DM Dataset that contains eight weeks' data for 6 patients to the 2018 Intern Health Study with over 1000 subjects \citep[see e.g.,][]{necamp2020assessing}. It is also applicable to data generated from online video games where both $N$ and $T$ are allowed to grow to infinity.

Empirically, we show that our procedure outperforms  existing learning algorithms using both synthetic datasets and a real dataset from the mobile health application. %and the other from a technological company. %Specifically, for each of the Q-learning method, we construct the corresponding pseudo outcomes to derive the estimated optimal policy. We find that the value under the resulting policy get improved after applying our de-biasing procedure in almost all cases. In addition, 
We remark that most papers in the existing literature use synthetic datasets to evaluate the performance of different RL algorithms. %It remains largely unknown how these algorithms perform in real-world domains. 
Results in our paper offer a useful evaluation tool for assessing these algorithms in real applications. 

The rest of this article is organised as follows. In Section \ref{sec:preliminary}, we introduce some basic concepts in RL, describe the data generating process and formulate the problem. In Section \ref{sec:QvsA}, we demonstrate the advantage of A-learning over Q-learning by comparing their rate of convergence. The proposed algorithm is formally presented in Section \ref{sec:PEAL}. In Section \ref{sec:theory}, we study the statistical properties of our algorithm, proving that our estimated policy achieves a faster rate of convergence than existing Q-learning type algorithms. In Section \ref{sec:syndata}, we investigate the finite sample performance of the proposed algorithm using Monte Carlo simulations. In Section \ref{sec:mobilehealth}, we use the OhioT1DM Dataset to further demonstrate the empirical advantage of the proposed algorithm over other baseline algorithms. Finally, we conclude our paper by a discussion section. Proofs of our major theorems are presented in Section \ref{sec:proof} of the supplementary article.  

\section{Preliminaries}\label{sec:preliminary}
%This section is organised as follows. 
We first formulate the policy optimization problem in infinite horizon settings. We next briefly review Q-learning. 
%introduce the notion of the optimal Q-function and optimal contrast function. %Finally, we discuss the minimax-optimal statistical convergence rate in supervised learning, as it is closely related to our proposal.
\subsection{Problem Formulation}\label{sec:MDP}
%We consider an offline setting for policy optimization. Consi
RL is concerned with solving sequential decision making problems in an unknown environment. The observed data can be summarized into a sequence of state-action-reward triplets over time. At each time $t\ge 0$, the decision maker observes some features from the environment, summarized into a {\it state} vector $S_t\in \mathbb{S}$ where the state space $\mathbb{S}$ is assumed to be a subset of $\mathbb{R}^d$. The decision maker then selects an {\it action} $A_t$ from the action space $\mathbb{A}$. The environment responds by providing the decision maker with an immediate {\it reward} $R_t\in \mathbb{R}$ and moving to the next state $S_{t+1}$. In this paper, we focus on the setting where $\mathbb{A}$ is discrete. Extensions to the continuous action space are discussed in Sections \ref{sec:actdis} and \ref{sec:kernel}. The state space $\mathbb{S}$ can be either continuous or discrete. 

A policy defines the agent's way of behaving. A {\it history-dependent} policy $\pi$ is a sequence of decision rules $\{\pi_t\}_{t\ge 0}$ where each $\pi_t$ is a function that maps the observed data history to a probability distribution function on the action space at time $t$. When these decision rules are time-homogeneous (i.e., $\pi_1=\pi_2=\cdots=\pi_t=\cdots$) and depend on the past data history only through the current state vector, the resulting policy is referred to as a {\it stationary} policy. Following $\pi$, the discounted cumulative reward that the decision maker receives is referred to as the {\it value} function, 
\begin{eqnarray*}
	V^{\pi}(s)=\sum_{t=0}^{+\infty} \gamma^t \Mean^{\pi} (R_t|S_0=s),
\end{eqnarray*}
where the expectation $\Mean^{\pi}$ is taken by assuming that actions are assigned according to $\pi$ and $0\le \gamma<1$ is a discounted factor that balances the long-term and short-term rewards. The objective of policy optimization is to identify an optimal policy $\pi^{\tiny{opt}}$ that maximizes the value, i.e., $\pi^{\tiny{opt}}=\argmax_{\pi} \Mean V^{\pi}(S_0)$.  
%The objective of policy optimization is to identify an optimal policy that maximizes the following discounted cumulative reward, 

We model the data generating process by a Markov decision process \citep[MDP,][]{Puterman1994}. Specifically, we impose the following Markov assumption (MA) and conditional mean independence assumption (CMIA). 
%Formally speaking, it is a function that maps the observed data history to a probability distribution function at each time. 
%. Suppose there is an agent that tries to learn and interact with certain environment. At each point, 

\smallskip

\noindent (MA) There exists some function $q$ such that for any $t\ge 0$, $\mathcal{S}\in \mathbb{S}$, we have
\begin{eqnarray*}
	\prob(S_{t+1}\in \mathcal{S}|\{S_{j},A_{j},R_{j}\}_{0\le j\le t})=\int_{\mathcal{S}}q(s;A_{t},S_t)ds.
\end{eqnarray*}

\noindent (CMIA) There exists some reward function $r$ such that for any $t\ge 0$, we have
\begin{eqnarray*}
	\Mean(R_{t}|S_{t},A_{t},\{S_{j},A_{j},R_{j}\}_{0\le j< t})=r(A_t,S_t).
\end{eqnarray*}

We make a few remarks. First, MA requires the future state to be conditional independent of the past data history given the current state-action pair. The function $q$ corresponds to the Markov transition density function that characterizes the state transitions. This assumption is testable from the observed data \citep[see e.g.,][]{shi2020does}. Second, under MA, CMIA is automatically satisfied when $R_t$ is a deterministic function of $S_t,A_t$ and $S_{t+1}$. The latter assumption is commonly imposed in the literature \citep{Ertefaie2018,luckett2019}. CMIA is weaker than this assumption. 

Second, these two assumptions lay the foundations of the existing state-of-the-art RL algorithms (e.g., DQN). Specifically, they guarantee the existence of an optimal stationary policy that is no worse than any history-dependent policies \citep[see e.g.,][]{Puterman1994}. It allows us to restrict our attentions to the class of stationary policies. For any such policy $\pi$, we use $\pi(\bullet|s)$ to denote the probability mass function that the decision maker will follow to select actions given that the environment is in the state $s$. 

The observed data consist of $N$ trajectories. Specifically, let $\{(S_{i,t},A_{i,t},R_{i,t},S_{i,t+1})\}_{0\le t< T}$ be the data collected  from the $i$-th trajectory where $T$ is the termination time. We assume these trajectories are independent copies of $\{(S_t,A_t,R_t)\}_{t\ge 0}$. Our objective is to learn $\pi^{\scriptsize{opt}}$ based on this offline dataset. 

\subsection{Q-learning}\label{sec:optQcon}
For a given policy $\pi$, we define the state-action value function (better known as the Q-function) under $\pi$ as 
\begin{eqnarray*}
	Q^{\pi}(a,s)=\sum_{t\ge 0} \gamma^t \Mean^{\pi} (R_{t}|A_{0}=a,S_{0}=s).
\end{eqnarray*}
It represents the average cumulative reward that the decision maker will receive if they select the action $a$ initially and follow $\pi$ afterwards. In addition, notice that
\begin{eqnarray}\label{eqn:bellmanpi}
	\begin{split}
		Q^{\pi}(a,s)=\Mean (R_0|A_{0}=a,S_{0}=s)+\gamma \left\{\sum_{t\ge 0} \gamma^t\Mean^{\pi} (R_{t+1}|A_{0}=a,S_{0}=s)\right\}\\
		=r(a,s)+\gamma \left\{\sum_{t\ge 0} \gamma^t \Mean^{\pi}[\Mean^{\pi} (R_{t+1}|A_1,S_1,A_{0}=a,S_{0}=s)|A_0,S_0]\right\}\\
		=r(a,s)+\gamma \Mean^{\pi} \{Q^{\pi}(A_1,S_1)|A_0=a,S_0=s\}\\=r(a,s)+\gamma \int_{s'} \sum_{a'}\pi(a'|s')Q(a',s')q(s';a,s)ds',
	\end{split}	
\end{eqnarray}
where the third equation follows from CMIA and the definition of $Q^{\pi}$ and the last equation follows from MA. The above equation is referred to as the Bellman equation for $Q^{\pi}$.

Define optimal Q-function $Q^{\tiny{opt}}$ as $Q^{\tiny{opt}}(a,s)=\max_{\pi} Q^{\pi}(a,s)$ for any state-action pair $(a,s)$. Under MA and CMIA, it can be shown that $\pi^{\scriptsize{opt}}$ satisfies 
\begin{eqnarray}\label{eqn:optpi}
	\pi^{\scriptsize{opt}}(a|s)=\mathbb{I}\left\{a=\argmax_{a'} Q^{\tiny{opt}}(a',s)\right\},\,\,\,\forall a,s,
\end{eqnarray}
where %sargmax denotes the smallest maximizer when the argmax is not unique and 
$\mathbb{I}\{\cdot\}$ denotes the indicator function. In addition, we have $Q^{\tiny{opt}}=Q^{\pi^{\tiny{opt}}}$. Similar to \eqref{eqn:bellmanpi}, one can show that $Q^{\tiny{opt}}$ satisfies the following Bellman optimality equation,
\begin{eqnarray}\label{Bellopt}
	Q^{\tiny{opt}}(a,s)=r(a,s)
	+\gamma \int_{s'} \max_{a'} Q^{\tiny{opt}}(s',a')q(s';a,s)ds',
\end{eqnarray} 
or equivalently,
\begin{eqnarray}\label{Bellopt1}
	Q^{\tiny{opt}}(A_t,S_t)=\Mean \left\{R_t+\gamma \max_{a}Q^{\tiny{opt}}(a,S_{t+1})|A_t,S_t\right\}.
\end{eqnarray} 
Equations \eqref{eqn:optpi} and \eqref{Bellopt1} form the basis for all Q-learning type algorithms. Specifically, these algorithms first estimate the optimal Q-function by solving \eqref{Bellopt1} and then derive the estimated optimal policy based on \eqref{eqn:optpi}. Take the fitted Q-iteration algorithm as an example. It iteratively updates the optimal Q-function using supervised learning. At each iteration, the input includes $(A_t,S_t)$ that serves as the ``predictors" and $R_t+\gamma \max_{a}\widetilde{Q}(a,S_{t+1})$ that serves as the ``response" where $\widetilde{Q}$ denotes the current estimate of the optimal Q-function. 
%For instance, the fitted Q-iteration algorithm iteratively updates the optimal Q-function 

Finally, we introduce the contrast function. %The idea of separating Q-functions into value and advantage functions originates from . 
For a given $\pi$, define the contrast function associated with $\pi$ as $\tau^{\pi}(a,s)=Q^{\pi}(a,s)-Q^{\pi}(a_0,s)$\footnote{Here, we define the contrast function as the difference between two Q-functions. Alternatively, one may define $\tau^{\pi}$ to be the advantage function, i.e., the difference between $Q^{\pi}$ and $V^{\pi}$.} for some $a_0\in \mathbb{A}$. In practice, the control arm $a_0$ could be set to the action that occurs the most in the data. \change{This is because the baseline Q-function $Q^{\pi}(a_0,s)$ needs to be accurately estimated in order to consistently estimate the contrast function. Hence, it is natural to consider the most frequently selected arm, which has the largest number of observations to learn the baseline Q-function.} Let $\tau^{\tiny{opt}}(a,s)=\tau^{\pi^{\tiny{opt}}}(a,s)$ be the optimal contrast function. Similar to \eqref{eqn:optpi}, we obtain that
%
%\begin{eqnarray}\label{eqn:optpiA}
$$\pi^{\scriptsize{opt}}(a|s)=\mathbb{I}\{a=\argmax_{a'} \tau^{\tiny{opt}}(a',s)\},$$ for any $a$ and $s$. 
%\end{eqnarray}
Consequently, to estimate the optimal policy, it suffices to estimate $\tau^{\tiny{opt}}$. This observation motivates the proposed advantage learning method. 

%{\color{red}Throughout the paper, we assume the }
%We present our estimating procedure in the next section. 
%To conclude this section, we introduce the following stationarity assumption (SA). %on the data generating process. 

%Since the data is generated offline, we impose the following stationarity assumption (SA) that is commonly imposed in the literature on off-policy evaluation.
%
%(SA)(i) Actions are generated by a stationary policy $b$. (ii) The process $\{S_{t}\}_{t\ge 0}$ is strictly stationary.
%
%The policy $b$ in (SA) is referred to as the behavior policy. We assume $b$ is known is bounded away from $0$. When $b$ is unknown, we can apply state-of-the-art supervised learning algorithms to the observed data to estimate $b$. Consequently, our results will be the same without this restriction. Under (SA)(i), the process $\{S_{t}\}_{t\ge 0}$ forms a time-homogeneous Markov chain, and thus  (SA)(ii) is  automatically satisfied when the initial distribution of $\{S_{t}\}_{t\ge 0}$ equals its stationary distribution. Without special saying, we assume MA, CMIA and SA hold throughout this paper. 
\section{Q- v.s. A-learning}\label{sec:QvsA}
This section is organised as follows. We first introduce the minimax-optimal statistical convergence rate in supervised learning, which serves as an evaluation metric to compare various supervised learning algorithms. We next demonstrate the advantage of A-learning over Q-learning by comparing the \change{worst-case} convergence rates of the estimated optimal contrast and Q-functions. Finally, we discuss the challenge of developing statistically efficient A-learning algorithms. 
%next discuss the limitation of directly applying the Bellman's optimality equation to update the contrast function.
\subsection{Minimax optimal statistical convergence rate}
Consider a supervised learning setup where we have given i.i.d. random vectors $\{(X_i,Y_i):1\le i\le n\}$. Our objective is to predict the value of the response $Y$ from the value of the feature $X\in \mathbb{S}$. The aim is to construct a best predictor to approximate the conditional mean function $m(X)=\Mean (Y|X)$. For any such predictor $\widehat{m}$, its prediction accuracy is measured by the root mean square error,
\begin{eqnarray}\label{eqn:square}
	\sqrt{\Mean |\widehat{m}(X)-m(X)|^2}.
\end{eqnarray}

Suppose $m$ belongs to the class of $p$-smooth (also known as H{\"o}lder smooth with exponent $p$) functions. When $p$ is an integer, 
%on $\mathbb{S}$ (see Appendix \ref{sec:moresmooth} for detailed definitions) for some $p>0$. When $p$ is an integer, this condition 
this condition essentially requires $m$ to have bounded derivatives up to the $p$th order. Formally speaking, for a $J$-tuple $\alpha=(\alpha_1,\dots,\alpha_J)^{\top}$ of nonnegative integers and a given function $h$ on $\mathbb{S}$, let $D^{\alpha}$ denote the differential operator:
\begin{eqnarray*}
	D^{\alpha}h(s)=\frac{\partial^{\|\alpha\|_1} h(s)}{\partial s_1^{\alpha_1}\cdots\partial s_J^{\alpha_J}}.
\end{eqnarray*}
Here, $s_j$ denotes the $j$th element of $s$. For any $p>0$, let $\floor{p}$ denote the largest integer that is smaller than $p$. The class of $p$-smooth functions is defined as follows:
\begin{eqnarray*}
	\Lambda(p,c)=\left\{h:\sup_{\|\alpha\|_1\le \floor{p}} \sup_{s\in \mathbb{S}} |D^{\alpha} h(s)|\le c, \sup_{\|\alpha\|_1=\floor{p}} \sup_{\substack{s_1,s_2\in \mathbb{S}\\ s_1\neq s_2}} \frac{|D^{\alpha} h(s_1)-D^{\alpha} h(s_2)|}{\|s_1-s_2\|_2^{p-\floor{p}}}\le c \right\},
\end{eqnarray*}
for some constant $c>0$. 
When $0<p\le 1$, we have $\floor{p}=0$. It is equivalent to require $h$ to satisfy $\sup_{s_1,s_2} |h(s_1)-h(s_2)|/\|s_1-s_2\|_2^p\le c$. The notion of $p$-smoothness is thus reduced to the H{\"o}lder continuity.

\citet{stone1982optimal} showed that the optimal minimax rate of convergence for $\widehat{m}$ is given by
\begin{eqnarray}\label{eqn:rate}
	n^{-p/(2p+d)},
\end{eqnarray}
where $d$ denotes the dimension of $\mathbb{S}$.  In other words, for any data-dependent predictor $\widehat{m}$, there exists some $p$-smooth function $m$ such that \eqref{eqn:square} decays at a rate of \eqref{eqn:rate}. This rate cannot be improved unless imposing certain parametric model assumptions on $m$. Notice that \eqref{eqn:rate} increases with the smoothness parameter $p$. In other words, the smoother the underlying regression function, the faster \change{worst-case} rate of convergence a supervised learner could achieve. 

\change{Finally, we remark that we focus on the class of H{\"o}lder smooth functions throughout this paper. Alternatively, one may consider the Sobolev space. Discussion of Sobolev and H{\"o}lder spaces can be found in \cite{gine2021mathematical}.}
%
%Finally, we present the proposed method. 
\subsection{Modelling contrast or Q-function?}\label{sec:conQ}
%To demonstrate the advantage of A-learning over Q-learning, we focus on comparing the best possible convergence rates of estimated optimal contrast and Q-functions. 
We assume the state space $\mathbb{S}$ is continuous and both the transition function $q(s';a,\bullet)$ and reward function $r(a,\bullet)$ belong to the class of $p$-smooth functions on $\mathbb{S}$ for some $p>0$.  \change{The p-smoothness assumption is likely to hold in many mobile health applications and we delegate the related discussions in Section \ref{subsec:psmooth}. }Under this condition, the optimal Q-function is $p$-smooth as well \citep[see Section 4,][]{fan2020theoretical}. %Let $d$ denote the dimension of $\mathbb{S}$, 
\citet{fan2020theoretical} proved that the Q-function computed by DQN achieves a rate of $(NT)^{-p/(2p+d)}$ up to some logarithmic factors. As they commented, this rate achieves the minimax-optimal statistical convergence rate in \eqref{eqn:rate} within the class of $p$-smooth functions and cannot be further improved. 

Since the optimal contrast function corresponds to the difference between two optimal Q-functions, $\tau^{\tiny{opt}}$ is at least at smooth as $Q^{\tiny{opt}}$. On the other hand, there are cases where $\tau^{\tiny{opt}}$ is strictly ``smoother" than $Q^{\tiny{opt}}$, leading to a possibly faster \change{worst-case} rate of convergence according to the minimax-optimal rate formula. %Let $\Lambda(p)$ denote the class of $p$-smooth functions. 
We consider two examples to elaborate. 
\begin{example}[Independent Transitions]\label{exam:ind}
	Consider the setting where the state transitions are independent, i.e., $q(s';a,s)=q(s')$ is independent of $(a,s)$. Then $Q^{\tiny{opt}}(a,s)=r(a,s)+C$ for some constant $C>0$ that is independent of $s$ and $a$. Suppose the reward function has the following decomposition 
	\begin{eqnarray*}
		r(a,s)=r^*(a,s)+r_0(s), 
	\end{eqnarray*}
	for some $p$-smooth baseline reward function $r_0$ and $p^*$-smooth function $r^*$ with $p^*>p$. It follows that $Q^{\tiny{opt}}(a,\bullet)$ is $p$-smooth whereas $\tau^{\tiny{opt}}(a,\bullet)=r^*(a,\bullet)-r^*(a_0,\bullet)$ is $p^*$-smooth.
\end{example}
%\textbf{Example 1 (). }

\begin{example}[Dependent Transitions]\label{exam:d}
	Suppose $q$ has the following decomposition 
	\begin{eqnarray}\label{eqn:qdecomp}
		q(s';a,s)=q^*(s';a,s)+q_0(s';s),
	\end{eqnarray}
	where $q^*(s';\bullet,a)$ has derivatives up to the $p^*$-th order whereas $q_0(s';\bullet)$ has derivatives up to the $p$-th order with $p<p^*$. By changing the order of integration and differentiation with respect to $s$, we can show that the second term on the right-hand-side (RHS) of \eqref{Bellopt} is $p$-smooth. Suppose $r(a,\bullet)$ has derivatives of all orders. It follows from \eqref{Bellopt} that $Q^{\tiny{opt}}$ is $p$-smooth. 
	
	On the contrary, by \eqref{Bellopt} and \eqref{eqn:qdecomp}, we have that
	\begin{eqnarray*}
		\tau^{\tiny{opt}}(a,s)=r(a,s)-r(a_0,s)
		+\gamma \int_{s'} \max_{a'} Q^{\tiny{opt}}(a',s')q^*(s';a,s)ds'.
	\end{eqnarray*}
	Using similar arguments, we can show that the last term on the RHS is $p^*$-smooth. This in turn implies that $\tau^{\tiny{opt}}$ is $p^*$-smooth as well.
	%we can show $\tau^{\tiny{opt}}$ is $p^*$-smooth. 
\end{example}
\change{To conclude this section, we remark that the minimax rate for the contrast function has been recently established in singe-stage decision making \citep{kennedy2022minimax}. In infinite horizon settings with tabular models, several papers have investigated the minimax-optimality of the Q-learning estimator \citep[see e.g.,][]{wainwright2019variance,li2020sample,li2021q}. In settings with continuous state space, a recent proposal of \citet{chen2022well} derived a minimax lower bound for the Q-function estimator under a fixed target policy and found that the rate matches those for nonparametric regression \citep{stone1982optimal}. We expect that similar arguments can be applied to formally obtain the minimax lower bounds for the estimated optimal Q- or contrast function.}
%\textbf{Example 2 (). }
\subsection{The challenge}\label{sec:updatecontrast}
So far we have shown that the \change{worse-case} convergence rate of the estimated optimal contrast function is faster than that of the estimated optimal Q-function. However, it remains challenging to devise an advantage learning algorithm that achieves such a rate of convergence. To elaborate, let us revisit the Bellman optimality equation in \eqref{Bellopt1}. By the definition of the optimal contrast function, it follows that
%it is possible to devise an algorithm that is more statistically efficient than Q-learning by directly modelling the optimal contrast function. 
%
%In this section, we discuss the limitation of directly applying the Bellman's optimality equation to update $\tau^{\tiny{opt}}$. Specifically, it follows from \eqref{Bellopt} that 
%\begin{eqnarray*}
%	Q^{\tiny{opt}}(S_{t},A_{t})=\Mean \left\{\left.R_{i,t}+\gamma \max_{a} Q^{\tiny{opt}}(S_{t+1},a)\right|S_{t},A_{t}\right\},
%\end{eqnarray*}
%for any $t$. As such, we have
\begin{eqnarray}\label{eqn:bellman}\quad
	\tau^{\tiny{opt}}(A_t,S_t)=\Mean \left\{R_{t}+\gamma \max_{a} \tau^{\tiny{opt}}(a,S_{t+1})\left.+\gamma Q^{\tiny{opt}}(a_0,S_{t+1})\right|A_t,S_t\right\}-Q^{\tiny{opt}}(a_0,S_t).
\end{eqnarray}
The presence of the nuisance function $Q^{\tiny{opt}}(a_0,\bullet)$ in the above equation poses a serious challenge to efficient estimation of $\tau^{\tiny{opt}}$. A simple solution is to apply Q-learning type algorithms to learn the nuisance function, plug in this estimator in \eqref{eqn:bellman} and update $\tau^{\tiny{opt}}$ using e.g., fitted Q-iteration. 
%%The above equation involves both $\tau^$
%
%could be applied to simultaneously update $\tau^{\tiny{opt}}$ and $Q^{\tiny{opt}}(\bullet,a_0)$ based on the observed data. 
However, such an approach would yield a sub-optimal solution. 
This is because the estimation error of the initial Q-estimator would directly affect that of the estimated contrast function. As a result, the estimated contrast would have the same convergence rate as the Q-estimator. %despite that $\tau^{\tiny{opt}}$ could be potentially ``smoother" than $Q^{\tiny{opt}}$. This implies that updating $\tau^{\tiny{opt}}$ according to \eqref{eqn:bellman} is not efficient.
%would yield a sub-optimal solution.
%To improve its rate of convergence, one needs to ensure that the contrast estimator 
%we derive a contrast estimator that is robust to the model misspecification of the Q-function. 
%To better illustrate the idea, we begin by considering estimating the advantage function $A^{\pi}$ under a given policy $\pi$. Later in this section we propose methods to estimate $A^{\tiny{opt}}$. The resulting estimated optimal policy can be derived based on \eqref{eqn:optpiA}. 
%\subsection{Learning the advantage function under a fixed policy $\pi$}
%We first present our full estimating procedure which consists of sample splitting, estimation of the Q-function and the density ratio (see the detailed definition below), construction of pseudo outcomes and supervised learning. We then discuss in detail some major steps of our algorithm. %Finally, we establish the oracle property of our estimator.
%We propose a supervised learning framework for reinforcement learning. 
%Unlike existing methods built upon Q-iterations, we construct pseudo outcomes that are asymptotically unbiased to the optimal Q-function. With these pseudo outcomes serving as the prediction target, the proposed method has a flavor of supervised learning, within an overall reinforcement learning framework. Therefore we name our method SupRL. 
\section{Statistically efficient A-learning}\label{sec:PEAL}
We first present the motivation of our algorithm. We next formally introduce our proposal. 
\subsection{A thought experiment}\label{sec:thought}
To illustrate the idea, in this section, let us consider a simplified model where the discounted factor $\gamma=0$ and the transitions are independent (see Example \ref{exam:ind}). In that case, we are interested in learning an optimal myopic policy the maximizes the short-term reward on average, which is essentially a single-stage decision making problem. By definition, the Q-function $Q^{\pi}$ and the contrast $\tau^{\pi}$ are independent of the policy $\pi$. Equation \eqref{eqn:bellman} can be rewritten as 
\begin{eqnarray}\label{eqn:tausinglestage}
	\tau(A_t,S_t)=\Mean (R_t|A_t,S_t)-Q(a_0,S_t)
\end{eqnarray}
where $Q(a_0,s)=\Mean (R_t|A_t=a_0,S_t=s)$. 

A-learning algorithms developed in the statistics literature can be employed to learn the contract function in this setting. They are motivated by the following identity,
\begin{eqnarray}\label{eqn:tausinglestage1}
	\sum_{a\neq a_0} \Mean [\{\mathbb{I}(A_t=a)-\prob(A_t=a|S_t)\}\{R_t- \tau(A_t,S_t)-Q(a_0,S_t) \}|S_t]=0.
\end{eqnarray}
Unlike Equation \eqref{eqn:tausinglestage}, the above equation is doubly-robust. It holds when either the propensity score $\prob(A_t=\bullet|S_t)$ or the Q-function $Q(a_0,\bullet)$ is correctly specified. This motives the following two-step procedure. In the first step, we first estimate the propensity score and the Q-function from the observed data. In the second step, we plug in these estimates in \eqref{eqn:tausinglestage1} to estimate the contrast function. Such a two-step method guarantees the estimated contrast to be robust to the potential model misspecification of the Q-function. 

When linear sieves are used to approximate $\tau$, i.e., $\tau(a,s)=\phi(a,s)^\top \beta_0$ for some basis function $\phi$, an estimating equation for $\beta_0$ can be constructed based on \eqref{eqn:tausinglestage1}. A Dantzig selector-type regularization can be applied when the number of basis functions is large \citep{shi2018high}. To employ more flexible machine learning methods, we can consider the following least-square objective function,% \citep[see e.g.,][]{lu2013variable}, 
\begin{eqnarray*}
	\sum_{\substack{i,t\\a\neq a_0}} \Big[\underbrace{\left\{ \frac{\mathbb{I}(A_{i,t}=a)}{\prob(A_{i,t}=a|S_{i,t})}- \frac{\mathbb{I}(A_{i,t}=a_0)}{\prob(A_{i,t}=a_0|S_t)}\right\}\{R_{i,t}-Q(A_{i,t},S_{i,t})\}+Q(a,S_{i,t})-Q(a_0,S_{i,t})}_{\psi(S_{i,t},A_{i,t},R_{i,t},a)}\\
	-\tau(a,S_{i,t})\Big]^2.
\end{eqnarray*}
Here, $\psi(S_{i,t},A_{i,t},R_{i,t},a)$ serves as a pseudo outcome for $\tau(a,S_{i,t})$. It is derived based on augmented inverse probability weighting \citep[AIPW, see e.g.,][]{bang2005doubly}. One can similarly show that $\Mean \{\psi(S_{i,t},A_{i,t},R_{i,t},a)|S_{i,t}\}$ is unbiased to $\tau(a,S_{i,t})$ when either the propensity score or the Q-function is correctly specified. A by-product of the doubly-robustness property is that when both nuisance functions are estimated from the data, the bias of the pseudo outcome will converge at a faster rate than these estimated nuisance functions. This in turn allows the resulting estimated contrast to converge at a faster rate than the Q-function. See e.g., Section \ref{sec:theory} for details. %We remark that similar objective functions have been considered in the literature \citep[see e.g.,][]{lu2013variable,nie2017quasi,li2021robust}. 

Although the above solution is developed in single-stage decision making, the same principle can be applied to general sequential decision making problems in infinite horizons, as we detail in the next section. 

%To estimate the contrast function, one could first learn the propensity score and the Q-function from the observed data and then plug in these estimates in \eqref{eqn:tausinglestage1} 

\subsection{The complete algorithm}
Our proposal involves two key components. First, we apply existing off-policy evaluation methods to construct pseudo outcomes for the optimal contrast function. This effectively reduces the bias of the initial Q-estimators, as we show in Theorem \ref{thm:2} that the bias of our pseudo outcomes decays at  a much faster rate than initial Q-estimators. It in turn ensures that the estimated contrast is robust to the model misspecification of the Q-function, improving its rate of convergence. %We note that there are several OPE methods in the existing literature. Here, we consider the doubly-robust estimator proposed in \citet{kallus2019efficiently}. Meanwhile, other methods are equally applicable. 
Second, we learn $\tau^{\tiny{opt}}$ by directly minimizing the least square loss between the pseudo outcomes and the estimated contrast. This allows us to borrow the strength of supervised learning to improve the statistical efficiency for RL. We call this set of method SEAL --- short for {\it statistically efficient advantage learning}. 
%We refer to our method as SupRL. 

Our proposal consists of five steps, including data splitting, policy optimization, estimation of the density ratio, construction of pseudo outcomes,  and supervised learning. 
We  %present an overview of our algorithm and then 
next discuss each step in detail.  

\subsubsection{Step 1. Data splitting} We randomly divide the indices of all trajectories $\{1,\ldots,N\}$ into $\mathbb{L}$ subsets $\cup_{\ell=1}^{\mathbb{L}} \mathcal{I}_{\ell}$ with equal size, for some fixed integer $\mathbb{L}>0$. Let %$\mathcal{I}_{\ell}$ denote the indices of trajectories contained in the $\ell$-th data subset and 
$\mathcal{I}_{\ell}^c$ be the complement of $\mathcal{I}_{\ell}$. Data splitting allows us to use one part of the data $(\mathcal{I}_{\ell}^c)$ to train RL models and the remaining part $(\mathcal{I}_{\ell})$ to construct the pseudo outcomes. We could aggregate the estimate over different $\ell$ to get full efficiency. This allows the bias of the constructed pseudo outcomes to decay to zero under minimal conditions on the estimated RL models. We remark that data splitting has been commonly used in the statistics and machine learning literature \citep[see e.g.,][]{Cherno2018,romano2019,kallus2019efficiently}.

\subsubsection{Step 2. Policy optimization} For $\ell=1,\ldots,\mathbb{L}$, we apply existing state-of-the-art Q-learning type algorithms to the data subset in $\mathcal{I}_{\ell}^c$ to compute an initial Q-estimator $\widehat{Q}^{(\ell)}$ for $Q^{\tiny{opt}}$. Several algorithms can be applied here, as we elaborate below.
\begin{example}[DQN]\label{exam:DQN}
	The deep Q-network algorithm is a Q-learning type method that uses a neural network Q-function approximator and several tricks to mitigate instability. It was developed in online settings and shown to yield superior performance to previously known methods for playing Atari 2600 games. To handle offline data, at each time step, we sample a minibatch of transitions $\{(S_{i,t},A_{i,t},R_{i,t},S_{i,t+1})\}_{(i,t)\in \mathcal{M}}$ and update the parameter $\theta$ of the Q-network by the gradient of 
	\begin{eqnarray}\label{eqn:dqngradient}
		\sum_{(i,t)\in \mathcal{M}} \{R_{i,t}+\max_{a} \gamma Q_{\theta^*}(a,S_{i,t+1})-Q_{\theta}(A_{i,t},S_{i,t})\}^2,
	\end{eqnarray}
	where $Q_{\theta^*}$ is the target network whose parameter $\theta^*$ is updated every $T_{\tiny{\textrm{target}}}$ steps by letting $\theta^*=\theta$. In \cite{mnih2015human}, $T_{\tiny{\textrm{target}}}$ is set to 10000. As $T_{\tiny{\textrm{target}}}$ grows to infinity, performing $T_{\tiny{\textrm{target}}}$ stochastic gradient steps is equivalent to solve 
	\begin{eqnarray*}
		\argmin_{\theta} \sum_{i=1}^N\sum_{t=0}^{T-1} \{R_{i,t}+\max_{a} \gamma Q_{\theta^*}(a,S_{i,t+1})-Q_{\theta}(A_{i,t},S_{i,t})\}^2.
	\end{eqnarray*}
	In that sense, DQN shares similar spirits with the fitted Q-iteration algorithm \citep{fan2020theoretical}.
\end{example}

\begin{example}[Double DQN]
	The double DQN algorithm is very similar to DQN. It is developed to alleviate the overestimation bias of the learned Q-function. DQN is likely to overestimate the Q-function under certain conditions, due to the biased resulting from the maximization step $\max_a Q_{\theta^*}(a,S_{i,t+1})$ in \eqref{eqn:dqngradient}. See e.g., \cite{Sutton2018} for a detailed explanation of the maximization bias. To reduce this bias, it replaces the target $R_{i,t}+\max_{a} \gamma Q_{\theta^*}(a,S_{i,t+1})$ by 
	\begin{eqnarray*}
		R_{i,t}+\gamma Q_{\theta^*}(\argmax_a Q_{\theta}(a,S_{i,t+1}),S_{i,t+1}).
	\end{eqnarray*}
	In other words, it decomposes the maximization operation into action selection and state-action value evaluation, uses the Q-network for action selection and the target network for value evaluation. It was shown in \cite{van2016deep} that such a trick leads to much better performance on several games empirically. 
	%Double DQN employs sample-splitting to alleviate the bias. 
\end{example}

\begin{example}[Quantile DQN]\label{exam:QDQN}
	The quantile DQN algorithm can be viewed as a distributional version of DQN with quantile regression. Instead of directly learning $Q^{\tiny{opt}}$, the expected return given the initial state-action pair, it learns quantiles of the return based on the distributional analogue of Bellman's optimality equation \eqref{Bellopt1} and averages the learned quantiles to estimate $Q^{\tiny{opt}}$. Please refer to \cite{dabney2018distributional} for details. 
\end{example}

Given the Q-estimator $\widehat{Q}^{(\ell)}$, we denote the derived optimal policy (see Equation \ref{eqn:optpi}) by $\widehat{\pi}^{(\ell)}$, for $\ell=1,\cdots,\mathbb{L}$. %The optimal value $V^{\tiny{opt}}(s)$ can be estimated by $\widehat{V}^{(\ell)}(s)=\max_{a} \widehat{Q}^{(\ell)}(s,a)$.

%\textbf{Step 3. Estimation of the Density Ratio}. 
\subsubsection{Step 3. Estimation of the density ratio}
The purpose of this step is to learn a density ratio estimator based on each data subset. These estimators are further employed in the subsequent step to construct the pseudo outcomes for the optimal contrast function. 

We first define the density ratio. 
For a given policy $\pi$, let $p^{\pi}_t(\bullet,\bullet|a,s)$ denote the conditional probability density function of $(A_t,S_t)$ given the initial state-action pair $(a,s)$ assuming that the decision maker follows $\pi$ at time $1,2,\cdots,t$. We define the $\gamma$-discounted average visitation density as follows,
$$p^{\pi}_{\gamma}(\bullet,\bullet|a,s)=(1-\gamma) \sum_{t\ge 1} \gamma^{t-1} p^{\pi}_t(\bullet,\bullet|a,s).$$
Let $p_{\infty}(\bullet,\bullet)$ denote the density function of the limiting distribution of the stochastic process $\{(A_t,S_t)\}_{t\ge 0}$. %\change{With a slight abuse of notation, we will }
We define the density ratio as $$\omega^{\pi}(a',s'|a,s)=\frac{p^{\pi}_{\gamma}(a',s'|a,s)}{p_{\infty}(a',s')},$$ for any $s,a,s',a'$. Such a density ratio plays an important role in breaking the curse of horizon in off-policy evaluation\footnote{Notice that our defined density ratio is slightly different from those in the existing OPE literature in that it involves an initial state-action pair.} \citep{liu2018}. 

In this step, we learn the density ratio $\omega^{\widehat{\pi}^{(\ell)}}$ based on each data subset in $\mathcal{I}_{\ell}^c$, for $\ell=1,\ldots,\mathbb{L}$, where $\widehat{\pi}^{(\ell)}$ is the initial optimal policy computed in Step 2. Several methods can be used here, e.g., \citet{liu2018,uehara2019minimax,kallus2019efficiently}. In our implementation, we adopt the proposal in \citet{liu2018} to construct a mini-max loss function (see Equation \ref{optimize}) to estimate $\omega^{\widehat{\pi}^{(\ell)}}$. 
We use $\widehat{\omega}^{(\ell)}$ to denote the corresponding estimator. %We describe the proposed estimating procedure later in this section. 
Additional details are given in Appendix \ref{sec:weight} to save space.

\subsubsection{Step 4. Construction of Pseudo Outcomes} 
For $\ell=1,\cdots,\mathbb{L}$, consider a pair of indices $(i,t)$ with $i\in \mathcal{I}_{\ell}$, $0\le t <T$. 
In this step, we focus on constructing a pseudo outcome $\widetilde{Q}_{i,t,a}$ for $Q^{\tiny{opt}}(a,S_{i,t})$ for any $a\in \mathbb{A}$, based on the Q- and density ratio estimators computed in Steps 2 and 3. The corresponding pseudo outcome for $\tau^{\tiny{opt}}(a,S_{i,t})$ is given by $\widetilde{\tau}_{i,t,a}=\widetilde{Q}_{i,t,a}-\widetilde{Q}_{i,t,a_0}$.

To motivate our method, notice that by the Bellman equation, $$Q^{\tiny{opt}}(a,S_{i,t})=r(a,S_{i,t})+\gamma \Mean \{ V^{\pi^{\tiny{opt}}}(S_{i,t+1})|A_{i,t}=a,S_{i,t}\},$$ %where $V^{\pi^{\tiny{opt}}}$ is the state value under the optimal policy.  
%Therefore, 
it suffices to construct pseudo outcomes for $r(a,S_{i,t})$ and $\Mean \{V^{\pi^{\tiny{opt}}}(S_{i,t+1})|A_{i,t}=a,S_{i,t}\}$. 
%Suppose the actions are generated by a stationary behavior policy $b$. 
Pseudo outcomes for $r(a,S_{i,t})$ can be derived based on augmented inverse propensity-score weighting, as in Section \ref{sec:thought},
\begin{eqnarray*}
	\widetilde{r}_{i,t,a}=\widehat{r}^{(\ell)}(a,S_{i,t})+\frac{\mathbb{I}(A_{i,t}=a)}{\prob(A_{i,t}=a|S_{i,t})}\{R_{i,t}-\widehat{r}^{(\ell)}(a,S_{i,t})\},
\end{eqnarray*}
where $\widehat{r}^{(\ell)}$ denotes some estimator for the reward function $r$ computed using the data subset in $\mathcal{I}_{\ell}^c$. As we have commented, the use of AIPW ensures the unbiasedness of the pseudo outcome, 
%robustness of $\widetilde{r}_{i,t,a}$. Specifically, the above pseudo outcome is unbiased to $r(a,S_{i,t})$ 
regardless of whether $\widehat{r}^{(\ell)}$ is consistent to $r$ or not. 

As for $\Mean \{V^{\pi^{\tiny{opt}}}(a',S_{i,t+1})|A_{i,t}=a,S_{i,t}\}$, since $\pi^{\tiny{opt}}$ is unknown, we consider approximating it by 
\begin{eqnarray}\label{eqn:nu}
	\nu^{(\ell)}(a,S_{i,t})=\Mean \{V^{\widehat{\pi}^{(\ell)}}(S_{i,t+1})|S_{i,t},A_{i,t}=a\},
\end{eqnarray}
using the estimated optimal policy $\widehat{\pi}^{(\ell)}$. 

Suppose for now, the Markov transition density function $q$ is known. Then $\nu^{(\ell)}(S_{i,t},a)$ can be estimated using the existing policy evaluation methods. Here, we consider the doubly reinforcement learning method proposed by \citet{kallus2019efficiently}, 
\begin{eqnarray}\label{eqn:pseudooutcome0}
	\int_{s'}\max_{a'}\widehat{Q}^{(\ell)}(a',s')q(s';a,S_{i,t})ds'+\frac{1}{1-\gamma}\eta_{i,t,a},
\end{eqnarray}
where $\eta_{i,t,a}$ is an augmentation term, defined as 
\begin{eqnarray*}
	\frac{1}{|\mathcal{I}_{\ell}|T-1}\sum_{\substack{i'\in \mathcal{I}_{\ell}\\ (i',t')\neq (i,t) }} \widehat{\omega}^{(\ell)}(A_{i',t'},S_{i',t'}|a,S_{i,t})\{R_{i',t'}+\gamma \max_{a'}\widehat{Q}^{(\ell)}(a',S_{i',t'+1})-\widehat{Q}^{(\ell)}(A_{i',t'},S_{i',t'})\}.
\end{eqnarray*}
%where 
The second term $R_{i',t'}+\gamma \max_{a'}\widehat{Q}^{(\ell)}(a',S_{i',t'+1})-\widehat{Q}^{(\ell)}(A_{i',t'},S_{i',t'})$ denotes the Bellman residual constructed based on the initial Q-estimator. When the initial Q-estimator is consistent, it follows from the Bellman optimality equation that the mean of $\eta_{i,t,a}$ is asymptotically zero. The purpose of adding $\eta_{i,t,a}$ in \eqref{eqn:pseudooutcome0} is to offer additional protection against potential model misspecification of the initial Q-estimator. Specifically, it ensures that \eqref{eqn:pseudooutcome0} is unbiased to $\nu^{(\ell)}(S_{i,t},a)$ when either $\widehat{Q}^{(\ell)}$ or $\widehat{\omega}^{(\ell)}$ is consistent \citep[see e.g.,][]{kallus2019efficiently}. In addition, when the estimated ratio is consistent, it allows the bias of \eqref{eqn:pseudooutcome0} to decay to zero at a rate faster than $\widehat{Q}^{(\ell)}$. See Theorem \ref{thm:2} for a formal statement.
%is unbiased to $\nu^{(\ell)}(S_{i,t},a)$ even when the Q-estimator is biased. 
%when either the density ratio or the Q-estimator is consistent. %(see the proof of Theorem \ref{thm:1} for details). 

However, the pseudo outcome outlined in \eqref{eqn:pseudooutcome0} suffers from two major limitations. The first one is that the transition density $q$ is in general unknown in practice. The second one is that the calculation of $\eta_{i,t,a}$ requires $O(NT)$ number of flops, which is computationally intensive to implement on large datasets. 

Let $\widehat{\nu}^{(\ell)}$ be some estimator for $\nu^{(\ell)}$ (see Equation \ref{eqn:nu}) computed using $\{O_{i,t}\}_{0\le t< T_i, i\in \mathcal{I}_{\ell}^c}$. To address the first limitation, we again use augmented inverse probability weighting and replace the first term in \eqref{eqn:pseudooutcome0} by 
\begin{eqnarray*}
	\widetilde{\nu}_{i,t,a}=\widehat{\nu}^{(\ell)}(a,S_{i,t})+\frac{\mathbb{I}(A_{i,t}=a)}{\prob(A_{i,t}=a|S_{i,t})}\{\max_{a'}\widehat{Q}^{(\ell)}(a',S_{i,t+1})
	-\widehat{\nu}^{(\ell)}(a,S_{i,t}) \}. 
\end{eqnarray*}

Similar to $\widetilde{r}_{i,t,a}$, one can easily verify that $\widetilde{\nu}_{i,t,a}$ is unbiased to $\nu(a,S_{i,t})$ regardless of whether $\widehat{\nu}^{(\ell)}$ is consistent or not. To address the second limitation, we randomly sample a minibatch $\mathcal{M}_{i,t}$ from the set $\{(i',t'):(i',t')\neq (i,t),i'\in \mathcal{I}_{\ell},0\le t'<T\}$ to approximate $\eta_{i,t,a}$ by  $\widetilde{\eta}_{i,t,a}$, constructed based on the observations in $\mathcal{M}_{i,t}$ only. %In Appendix \ref{sec:moreoracle}, we detail the form for $\widetilde{\eta}_{i,t,a}$. 
Specifically, we define $\widetilde{\eta}_{i,t,a}$ by
\begin{eqnarray*}
	\frac{1}{|\mathcal{M}_{i,t}|}\sum_{(i',t')\in \mathcal{M}_{i,t}} \widehat{\omega}^{(\ell)}(A_{i',t'},S_{i',t'}|a,S_{i,t})\{R_{i',t'}+\gamma \max_{a'}\widehat{Q}^{(\ell)}(a',S_{i',t'+1})-\widehat{Q}^{(\ell)}(A_{i',t'},S_{i',t'})\},
\end{eqnarray*}
When $|\bullet|$ denotes the cardinality of a set. When $|\mathcal{M}_{i,t}|$ is much smaller than $NT$, it largely facilitates the computation.  
%where $|\mathcal{M}_{i,t}|$ denotes the batch size. 

Combining both parts yields the following, %pseudo outcome 
\begin{eqnarray*}
	&&\widetilde{r}_{i,t,a}+\gamma \widetilde{\nu}_{i,t,a}+\gamma(1-\gamma)^{-1}\widetilde{\eta}_{i,t,a}=\widehat{r}^{(\ell)}(a,S_{i,t})+\widehat{\nu}^{(\ell)}(a,S_{i,t})\\
	&+&\frac{\mathbb{I}(A_{i,t}=a)}{\prob(A_{i,t}=a|S_{i,t})}\{R_{i,t}+\max_{a'}\widehat{Q}^{(\ell)}(a',S_{i,t+1})-\widehat{r}^{(\ell)}(a,S_{i,t})
	-\widehat{\nu}^{(\ell)}(a,S_{i,t}) \}+\frac{\gamma}{1-\gamma}\widetilde{\eta}_{i,t,a}.
\end{eqnarray*}
%for $Q^{\tiny{opt}}(a,S_{i,t})$. 
Notice that $r(a,S_{i,t})+\gamma \nu^{(\ell)}(a,S_{i,t})=Q^{\widehat{\pi}^{(\ell)}}(a,S_{i,t})$ can be estimated by $\widehat{Q}^{(\ell)}(a,S_{i,t})$. Putting all the pieces together, we obtain the following pseudo outcome for $\widetilde{Q}_{i,t,a}$, defined by
\begin{eqnarray*}%\label{eqn:pseudooutcome}
	\begin{split}
		\widehat{Q}^{(\ell)}(a,S_{i,t})+\frac{\mathbb{I}(A_{i,t}=a)}{\prob(A_{i,t}=a|S_{i,t})}\{R_{i,t}+\gamma \max_{a'}\widehat{Q}^{(\ell)}(a',S_{i,t+1})-\widehat{Q}^{(\ell)}(A_{i,t},S_{i,t})\}+\frac{\gamma}{ (1-\gamma)} \widetilde{\eta}_{i,t,a}.
	\end{split}	
\end{eqnarray*}
As we have discussed, the pseudo outcome for the optimal contrast is obtained by $\widetilde{\tau}_{i,t,a}=\widetilde{Q}_{i,t,a}-\widetilde{Q}_{i,t,a_0}$. 

We again make some remarks. First, we employ cross-fitting to construct $\widetilde{\tau}_{i,t,a}$. That is, $\widehat{Q}^{(\ell)}$ and $\widehat{\omega}^{(\ell)}$ are computed by observations that are independent of $(S_{i,t},A_{i,t},R_{i,t},S_{i,t+1})$. This helps avoid overfitting which can easily result from the estimation of the Q-function and density ratio. Second, to simplify the presentation, we assume the propensity score is known. In practice, it can be estimated from the observed data and our theoretical results will be the same when the estimated propensity score satisfies certain rate of convergence. 
%. Such techniques have been commonly used in statistical inference; 
%\citep[see e.g.,][]{Cherno2018}. %Statistical properties of these pseudo outcomes are studied in the next section. 
%We next establish the statistical properties of $\widetilde{Q}_{i,t,a}$. 

\subsubsection{Supervised learning} In the final step, we factorize the contrast function $\tau^{\tiny{opt}}$ by some models $\tau\in \mathcal{T}$ and estimate the model parameter by minimizing the following objective function, 
\begin{eqnarray}\label{eqn:thetahat}
	\widehat{\tau}=\argmin_{\tau\in \mathcal{T}}\sum_{i=1}^N \sum_{t=0}^{T-1} \sum_{a\neq a_0} \{\widetilde{\tau}_{i,t,a}-\tau(a,S_{i,t})\}^2.
\end{eqnarray}
The corresponding estimated optimal policy is given by $\mathbb{I}\{a=\argmax_a^* \widehat{\tau}(a^*,s)\}$ for any $a$ and $s$. 

\begin{figure}[!t]
	\centering
	\includegraphics[width=6.5cm]{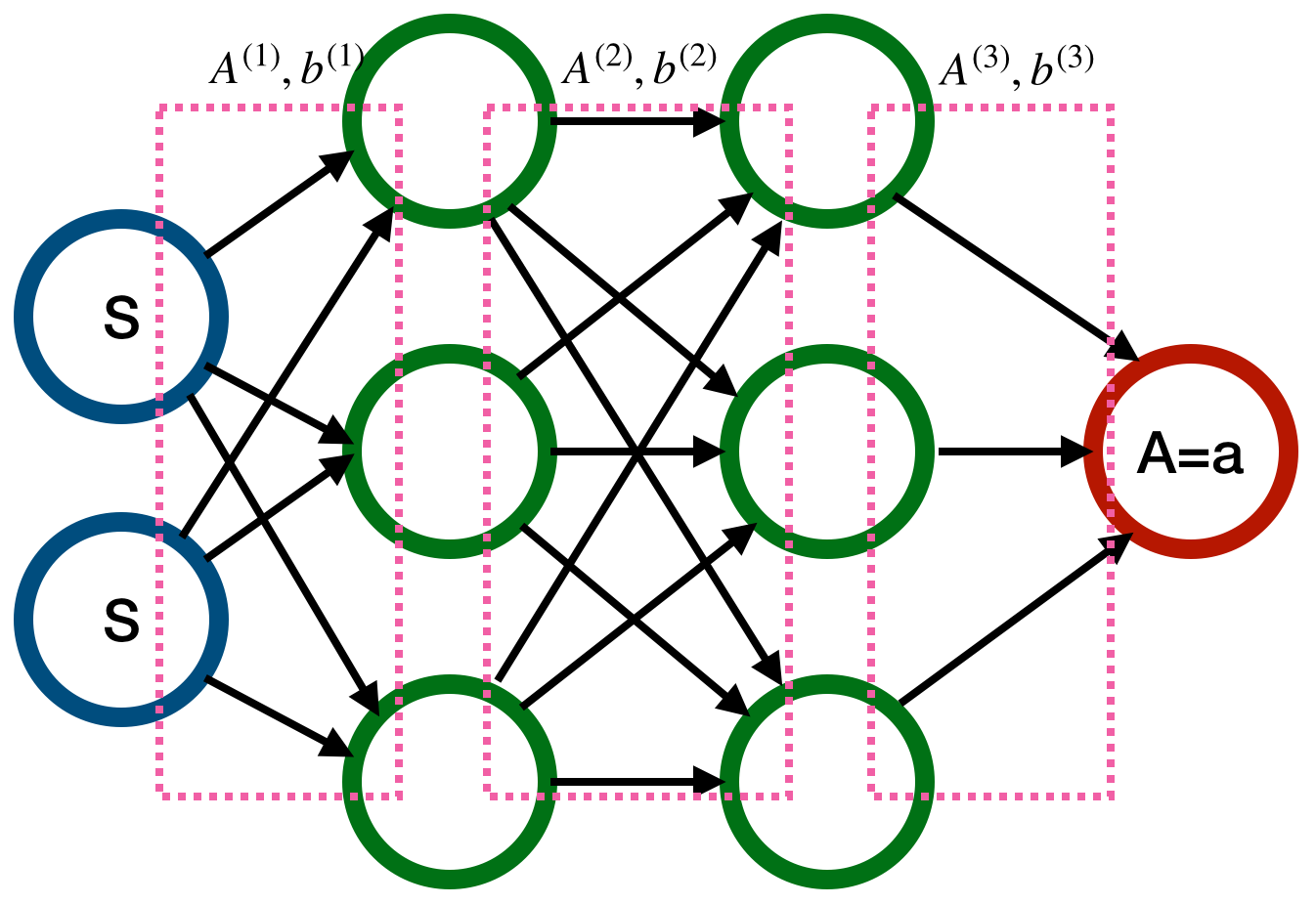}
	\caption{Illustration of a two-layer, fully connected DNN. The state is two-dimensional.}\label{fig1}
\end{figure}

To solve \eqref{eqn:thetahat}, it is equivalent to solve
\begin{eqnarray}\label{eqn:thetahat0}
	\argmin_{\tau\in \mathcal{T}_a}\sum_{i=1}^N \sum_{t=0}^{T-1} \{\widetilde{\tau}_{i,t,a}-\tau(a,S_{i,t})\}^2,
\end{eqnarray}
for each $a\neq a_0$. Many methods are available to solve \eqref{eqn:thetahat0}, as it is essentially a nonparametric regression problem. In our implementation, we set $\mathcal{T}_a$ to the class of deep neural networks (DNNs), so as to capture the complex dependence between the reward and the state-action pair. The input of the network is a $d$-dimensional vector, corresponding to the state (coloured in blue in Figure \ref{fig1}). The hidden units (coloured in green) are grouped in a sequence of $L_a$ layers. Each unit in the hidden layer is determined as a nonlinear transformation of a linear combination of the nodes from the previous layer. We use $W_a$ to denote the total number of parameters. These parameters are updated by the Adam algorithm \citep{Kingma2015}. 

%\begin{figure}
%	\centering
%	\includegraphics[width=4cm]{MLP1.png}
%	\caption{Illustration of DNN with two hidden layers, $m_0=2$, $m_1=m_2=3$. Here $u$ is the input, $A^{(\ell)}$ and $b^{(\ell)}$ denote the corresponding parameters to produce the linear transformation for the $(\ell-1)$th layer.}
%\end{figure}

%\vspace{-0.3cm}
\section{Theoretical findings}\label{sec:theory}%\vspace{-0.3cm}
%We briefly summarize our theoretical results to be presented in this section. In Theorem \ref{thm:1}, we show $\widetilde{Q}_{i,t,a}$ is asymptotically unbiased to $Q^{\widehat{\pi}^{(\ell)}}(S_{i,t},a)$ as either the estimated Q-function or density ratio is consistent. This demonstrate the doubly-robustness property of the constructed pseudo outcomes. 
We first summarize our theoretical findings. In Theorem \ref{thm:2}, %we derive the doubly-robustness property of our constructed pseudo outcomes. We show that  they are asymptotically unbiased to the Q-function under the derived optimal policy as either the initial Q-estimator or density ratio is consistent (see \eqref{condQ} and \eqref{condweight}). In Theorem \ref{thm:2}, 
we provide a finite sample bias analysis of the pseudo outcome, proving its bias decays at a faster rate than the initial Q-estimator. In Theorem \ref{thm:3}, we show our estimator for the optimal contrast achieves a faster rate of the convergence than the Q-estimator. In Theorem \ref{thm:4}, we show the resulting optimal policy achieves a larger value than those computed by Q-learning type algorithms. Finally, we discuss a potential limitation of the proposed method. All the error bounds derived in this paper  converge to zero when either $N$ or $T$ diverges to infinity. As commented in the introduction, this ensure our method is valid when applied to a wide range of real problems. 
\subsection{Finite sample bias analysis}%Let $\mathbb{F}_0$ denote the stationary distribution function of $\{S_{t}\}_{t\ge 0}$. 
In this section, we focus on deriving an error bound on the bias $\Mean (\widetilde{Q}_{i,t,a}|S_{i,t})-Q^{\tiny{opt}}(S_{i,t},a)$ as a function of the total number of observations $NT$.  %assuming both $\widehat{\omega}^{(\ell)}$ and $\widehat{Q}^{(\ell)}$ are consistent. 
We introduce the following conditions. 

\smallskip

\noindent (A1) The state space $\mathbb{S}$ is compact. There exists some constant $\alpha>0$ such that
\begin{eqnarray}\label{eqn:margin}
	\begin{split}
		%\lambda\left\{x\in \mathbb{X}:\max_{a} Q^{\tiny{opt}}(x,a)-\max_{a'\in \mathcal{A}-\argmax_{a} Q^{\tiny{opt}}(x,a)} Q^{\tiny{opt}}(x,a')\le \varepsilon\right\}=O(\varepsilon^{\alpha}),\\ \label{Gmargin}
		\lambda\left\{s\in \mathbb{S}:\max_{a} Q^{\tiny{opt}}(a,s)-\max_{a'\in \mathbb{A}-\argmax_{a} Q^{\tiny{opt}}(a,s)} Q^{\tiny{opt}}(a',s)\le \varepsilon\right\}=O(\varepsilon^{\alpha}),
	\end{split}
\end{eqnarray}
where $\lambda$ denotes the Lebesgue measure and the big-$O$ term in \eqref{eqn:margin} is uniform in $0<\varepsilon<\delta$ for some sufficiently small $\delta>0$. By convention, $\max_{a'\in \mathbb{A}-\argmax_{a} Q^{\tiny{opt}}(a,s)} Q^{\tiny{opt}}(a',s)=-\infty$ if the set $\mathbb{A}-\argmax_{a} Q^{\tiny{opt}}(a,s)$ is empty. 
%, and $\max_{a'\in \mathcal{A}-\argmax_{a} Q^{\tiny{opt}}(s,a)} Q^{\tiny{opt}}(s,a')=-\infty$ if the set $\mathcal{A}-\argmax_{a} Q^{\tiny{opt}}(s,a)=\emptyset$.  

\smallskip

\noindent (A2) $Q^{\tiny{opt}}(a,\cdot)$ is $p$-smooth and $\tau^{\tiny{opt}}(a,\cdot)$ is $p^*$-smooth for some $p^*<p$ and any $a$.

\smallskip

\noindent (A3) %There exist some constants $c_1,c_2,c_3>0$ such that for any $a$, $\ell$, the followings occur with probability tending to $1$, 
There exists some constant $0<c_0\le 1$ such that the followings hold for any $a$ and $\ell$, with probability approaching 1, 
\begin{eqnarray*}%\label{condQ}
	\Mean_{(a,s)\sim p_{\infty}} |\widehat{Q}^{(\ell)}(a,s)-Q^{\tiny{opt}}(a,s)|^2=O \{(NT)^{-2p/(2p+d)}\},\\
	\Mean_{(a,s)\sim p_{\infty}, (a',s')\sim p_{\infty}} |\widehat{\omega}^{(\ell)}(a',s'|a,s)-\omega^{\widehat{\pi}^{(\ell)}}(a',s'|a,s)|^2=O\{(NT)^{-c_0}\}.
\end{eqnarray*} 
%where the big-$O$ terms are uniform in $a$ and $\ell$.

\smallskip

\noindent (A4) The process $\{(S_{t},A_{t},R_{t})\}_{t\ge 0}$ is stationary and exponentially $\beta$-mixing \citep[see e.g.,][for detailed definitions]{Bradley2005}. 

\smallskip

\noindent (A5) The probability density function $p_{\infty}$ is uniformly bounded away from zero.

\smallskip 

In (A1), \change{we require the state space to be continuous. When it is discrete, we can replace the Lebesgue measure with the counting measure. Our theories are equally applicable.} we refer to the quantity $\max_{a} Q^{\tiny{opt}}(a,s)-\max_{a'\in \mathcal{A}-\argmax_{a} Q^{\tiny{opt}}(a,s)} Q^{\tiny{opt}}(                                                                                                                                                                                                                                                                                                                                                                                                                                                   a',s)$ as the ``margin" of the optimal Q-function. It measures the difference in value between $\pi^{\tiny{opt}}$ and the policy that assigns the best suboptimal treatment(s) at the first decision point and follows $\pi^{\tiny{opt}}$ subsequently. Such a margin-type condition is commonly used to bound the excess misclassification error \citep{Alex2004,Alex2007} and the regret of estimated optimal treatment regime \citep{qian2011,Alex2016,shi2020breaking}. %Under such a condition, 
Here, we impose Condition (A1) to bound the difference between $Q^{\widehat{\pi}^{(\ell)}}(S_{i,t},a)$ and $Q^{\tiny{opt}}(S_{i,t},a)$. %Specifically, under (A1), we show $Q^{\widehat{\pi}^{(\ell)}}(S_{i,t},a)$ will converge to $Q^{\tiny{opt}}(S_{i,t},a)$ at a much faster rate than the initial Q-estimator $\widehat{Q}^{(\ell)}(a,S_{i,t})$ (see the proof of Theorem \ref{thm:2} in the Appendix for details). 
This condition is mild. To elaborate, we consider a simple scenario where $\mathbb{A}=\{0,1\}$ and $a_0=0$. It follows that the margin equals $|\tau^{\tiny{opt}}(1,s)|$ if $\tau^{\tiny{opt}}(1,s)$ is nonzero and $+\infty$ otherwise. \eqref{eqn:margin} is thus equivalent to the following,
\begin{eqnarray}\label{eqn:margin1}
	\lambda\{s\in \mathbb{S}: 0<|\tau^{\tiny{opt}}(1,s)|\le \varepsilon\}=O(\varepsilon^{\alpha})
\end{eqnarray} 
The above condition can be satisfied in a wide range of settings. We consider three examples to illustrate. 
\begin{example}
	Suppose $\tau^{\tiny{opt}}(1,s)=0$ for any $s$. Then \eqref{eqn:margin1} is automatically satisfied. In this example, the two actions have the same effects. Any policy would achieve the same value. %function. 
	%It corresponds to a completely nonregular setting. 
\end{example}

\begin{example}
	Suppose $\inf_s |\tau^{\tiny{opt}}(1,s)|>0$. Then \eqref{eqn:margin1} is automatically satisfied for any sufficiently small $\varepsilon>0$. When the optimal contrast function is continuous, it requires $\tau^{\tiny{opt}}(1,s)$ to be always positive or negative as a function of $s$. As such, the optimal policy is nondynamic and will assign the same action at each time.
\end{example}

\begin{example}\label{exam1d}
	Consider the case where the state is one-dimensional. Suppose
	\begin{eqnarray*}
		\tau^{\tiny{opt}}(1,s)=\left\{\begin{array}{ll}
			s^{1/\alpha}, & \textrm{if}~s>0;\\
			0, & \textrm{otherwise},
		\end{array}
		\right.
	\end{eqnarray*}
	we have $\lambda\{s\in \mathbb{S}: 0<|\tau^{\tiny{opt}}(1,s)|\le \varepsilon\}=\lambda\{s\in \mathbb{S}: 0<s\le \varepsilon^{\alpha}\}=\varepsilon^{\alpha}$. \eqref{eqn:margin1} is thus satisfied.
\end{example}

In (A2), we assume the optimal contrast function is strictly ``smoother" than the optimal Q-function. As we have discussed, this assumption holds under several cases. See Examples \ref{exam:ind} and \ref{exam:d} in Section \ref{sec:conQ} for details.

In the first part of (A3), we assume the squared prediction loss of the estimated Q-function achieves a rate of $(NT)^{-2p/(2p+d)}$. As we have commented, this condition automatically holds when deep-Q network is used to fit the initial Q-estimator. The second part of (A3) is mild as the constant $c_0$ could be arbitrarily small. Suppose some parametric model (e.g., linear) is imposed to learn $\widehat{\omega}^{(\ell)}$. When the model is correctly specified, then we have $c_0=1$. When
kernels are used for function approximation, the rate $c_0$ can be established in a similar manner as in Theorem 5.4 of \citet{liao2020batch}. 
%and density ratio to satisfy certain convergence rates. This assumption is mild as $c_0$ is allowed to be an arbitrarily small constant. 

(A4) requires the $\beta$-mixing coefficients of the process $\{(S_{t},A_{t},R_{t})\}_{t\ge 0}$ to decay to zero at an exponential rate. These coefficients characterize the temporal dependence of the observations and are equal to zero when the data are independent. The smaller the coefficients, the weaker the dependence. When the propensity score is stationary over time, $\{(S_{t},A_{t},R_{t})\}_{t\ge 0}$ forms a time-homogeneous Markov chain. (A4) is automatically satisfied when the Markov chain is geometrically ergodic \citep[see Theorem 3.7 of][]{Bradley2005}. Geometric ergodicity is less restrictive than those  imposed in the existing reinforcement learning literature that requires observations to be  independent \citep[see e.g.,][]{degris2012off} or to follow a uniform-ergodic Markov chain \citep[see e.g.,][]{zou2019}. \change{We also remark that the stationarity assumption
	in (A1) is assumed only to simplify the technical proof. Our theoretical results are equally applicable even without this condition \citep[see e.g., the proof of Lemma 3 of][]{shi2020value}.} %convenience, since the Markov chain will eventually reach stationarity.
%These assumptions enable us to derive the following theorem. 
%These rates enable us to establish the following theorem. 
%nonasymptotic error bound on $\Mean [\widetilde{Q}_{i,t,a}|S_{i,t},\{O_{i,t}\}_{0\le t< T_i, i\in \mathcal{I}_{\ell}^c}]=Q^{\widehat{\pi}^{(\ell)}}(S_{i,t},a)$. 
%In (A2), we require the estimated Q-function and density ratio to satisfy certain uniform convergence rates. 

(A5) is very similar to the positivity assumption imposed in single-stage decision making. These assumptions enable us to derive the following theorem. 

\begin{thm}\label{thm:2}
	Assume (A1)-(A5) hold. $\widehat{Q}^{(\ell)}$, $\widehat{\omega}^{(\ell)}$ \change{and the rewards} are uniformly bounded. Then there exists some constant $\bar{c}>p/(2p+d)$ such that for any $a\in \mathbb{A}$,
	\begin{eqnarray*}
		\frac{1}{NT}\sum_{i,t}\Mean |\Mean (\widetilde{Q}_{i,t,a}|S_{i,t})-Q^{\tiny{opt}}(S_{i,t},a)|=O\{(NT)^{-\bar{c}}\}.
	\end{eqnarray*}
\end{thm}
Theorem \ref{thm:2} states that the conditional bias of $\widetilde{Q}_{i,t,a}$ decays at a rate of $(NT)^{-\bar{c}}$ on average. 
In comparison, under (A3), the squared prediction loss of the initial Q-estimator decays at a rate of $(NT)^{-2p/(2p+d)}$. Suppose the square bias and variance of $\widehat{Q}^{(\ell)}$ are of the same order. Then we expect $\Mean \{\widehat{Q}^{(\ell)}(a,S_{i,t})|S_{i,t}\}$ to approach $Q^{\tiny{opt}}(S_{i,t},a)$ at a rate of $(NT)^{-p/(2p+d)}$. Since $\bar{c}>p/(2p+d)$, biases of our pseudo outcomes are much smaller than the initial Q-estimators. %This in turn implies 

%Note that the 

\subsection{Efficiency enhancement} 
In this section, we establish the convergence rates of the estimated contrast function and the derived optimal policy. Without loss of generality, we assume the state space $\mathbb{S}=[0,1]^d$. We write $a_n\asymp b_n$ for two sequences $\{a_n\}_n$ and $\{b_n\}_n$ if there exists some universal constant $c\ge 1$ such that $c^{-1} a_n \le b_n\le ca_n$ for all $n$. 
\begin{thm}\label{thm:3}
	Assume the conditions in Theorem \ref{thm:2} hold. Then there exists DNN class $\{\mathcal{T}_a\}_a$ with $L_a\asymp \log(NT)$ and $W_a\asymp (NT)^{C}\log (NT)$ for some $C>d/(2p+d)$ and any $a\neq a_0$ such that with probability approaching $1$,
	%then under certain mild conditions on the DNN function class $\mathcal{T}$ (see Appendix \ref{sec:proof2} for details), there exists some constant $c_1>2p/(2p+d)$ such that 
	\begin{eqnarray*}
		\Mean_{s\sim p_{\infty}} |\widehat{\tau}(a,s)-\tau^{\tiny{opt}}(a,s)|^2=O\{(NT)^{-c_1}\},
	\end{eqnarray*}
	for some constant $2p/(2p+d)<c_1\le 2p^*/(2p^*+d)$, \change{where the expectation is taken with respect to the stationary state distribution}.
	%Further assume $\max_a \Mean |\widehat{\pi}^{(\ell)}(a|S_{0,0})-\pi^{\tiny{opt}}(a|S_{0,0})|^2=o(1)$ and $\sup_{s,a} \Mean |\omega^{\widehat{\pi}^{(\ell)}}(S_{0,0},s,a)-\omega^{\pi^{\tiny{opt}}}(S_{0,0},s,a)|^2=o(1)$. Then, both $\sqrt{NT}(\widehat{\theta}-\theta^*)$ and $\sqrt{NT}(\widehat{\theta}^*-\theta^*)$ are asymptotically normal. In addition, they have the same asymptotic variance. 
\end{thm}
Theorem \ref{thm:3} formally shows that our estimated contrast function converges at a faster rate than the Q-function computed by Q-learning type-estimators, leading to the desired efficiency enhancement property. \change{To illustrate why the estimated contrast converges faster, suppose we have access to some unbiased pseudo outcome for $\tau(a,S_{i,t})$. Then under (A2), the estimated contrast function would converge at a minimax optimal rate of $(NT)^{-p^*/(2p^*+d)}$, which is much faster than that of the Q-estimator. In practice, we do not have access to unbiased pseudo outcomes. As such, the rate would depend on the bias of the pseudo outcome $\widetilde{Q}_{i,t,a}-\widetilde{Q}_{i,t,a_0}$. Nonetheless, the efficiency enhancement property holds as long as the bias decays faster than the convergence rate of the Q-estimator. The latter assertion is confirmed in Theorem \ref{thm:2}.}

We next show this in turn leads to an improvement in the value. More specifically, for any policy $\pi$, define the integrated value function $\mathcal{V}(\pi)=\int_{\mathbb{S}} V^{\pi}(s)\nu_0(s)ds$ where $\nu_0$ denotes the density function of $S_0$. 
Let $\widehat{\pi}^{\tau}$ %and $\widehat{\pi}^{Q}$ 
denote the derived policies based on the estimated contrast function $\widehat{\tau}$. %and $\widehat{Q}$, respectively. 
\begin{thm}\label{thm:4}
	Assume %\eqref{eqn:Qrate} and 
	the conditions in Theorem \ref{thm:2} hold and  $\nu_0$ is uniformly bounded from above. Then
	\begin{eqnarray*}
		%		&&\mathcal{V}({\pi^{\tiny{opt}}})-\Mean \mathcal{V}(\widehat{\pi}^{Q})=O\{ (NT)^{-\alpha_0 p/(2p+d) } \},\\
		\mathcal{V}({\pi^{\tiny{opt}}})-\Mean \mathcal{V}(\widehat{\pi}^{\tau})=O\{ (NT)^{-\alpha_0 c_1/2} \},
	\end{eqnarray*}
	where $\alpha_0=(2+2\alpha)/(\alpha+2)>1$ and $\alpha$ is defined in (A1). 
\end{thm}
Let $\widehat{Q}$ denotes a Q-learning type estimator that satisfies
\begin{eqnarray}\label{eqn:Qrate}
	\Mean_{(a,s)\sim p_{\infty}} |\widehat{Q}(a,s)-Q^{\tiny{opt}}(a,s)|^2=O \{(NT)^{-2p/(2p+d)}\},
\end{eqnarray}
and $\widehat{\pi}^{Q}$ be the derived policy based on $\widehat{Q}$. Similar to Theorem \ref{thm:4}, we can show that $\Mean \mathcal{V}(\widehat{\pi}^{Q})$ converges at a rate of $\alpha_0 p/(2p+d)$. 
%Theorem \ref{thm:4} 
Based on the fact that $c_1>2p/(2p+d)$, it is clear that the value of our estimated policy converges to the optimal value at a faster rate than those of policies computed by Q-learning type algorithms. 

\change{The convergence rates in Theorems \ref{thm:3} and \ref{thm:4} relies crucially on the exponent $\alpha$ in the margin condition (A1) and the convergence rate of the estimated density ratio in (A3), i.e., $(NT)^{-c_0}$. The following corollary shows that under certain conditions on $\alpha$ and $c_0$, the exponent $c_1$ in both theorems achieve a maximum value of $2p^*/(2p^*+d)$. 
	
	\begin{coro}
		Suppose the conditions in Theorems \ref{thm:3} and \ref{thm:4} hold. Suppose $c_0\ge 2p^*/(2p^*+d)-2p/(2p+d)$ and $\alpha\ge 2[2-\{p^*/(2p^*+d)\}/\{p/(2p+d)\} ]^{-1}-2$. Then with proper choice of the DNN class $\{\mathcal{T}_a\}_a$, we have for any $a\neq a_0$ that
		\begin{eqnarray*}
			\Mean_{s\sim p_{\infty}} |\widehat{\tau}(a,s)-\tau^{\tiny{opt}}(a,s)|^2=O\{(NT)^{-2p^*/(2p^*+d)}\},
		\end{eqnarray*}
		with probability approaching $1$, and that  $\mathcal{V}({\pi^{\tiny{opt}}})-\Mean \mathcal{V}(\widehat{\pi}^{\tau})=O\{ (NT)^{-\alpha_0 p^*/(2p^*+d)} \}$. 
	\end{coro}
}

\change{Notice that we do not require the optimal policy to be unique in order to establish the regret bound of the estimated optimal policy. This is because our proposal is value-based which derives the optimal policy using the estimated advantage function. The advantage function is well-defined despite that the optimal policy might not be unique, and the regret bound decays to zero as long as the estimated advantage function is consistent. To elaborate, let us revisit Example \ref{exam1d}. By definition, when the state is nonpositive, both actions are optimal. The uniqueness assumption is thus violated. Nonetheless, the regret is zero since choosing either action is optimal.}%under the conditions in Theorem \ref{thm:2}, the advantage function can be consistently estimated, yielding a near-optimal policy whose value approaches the optimal value as the sample size diverges to infinity.}

Finally, we remark that although the proposed contrast function estimator converges at a faster rate than Q-learning type estimators, these rates are asymptotic. 
A potential limitation of the proposed method is that our estimated contrast function might have larger variance than Q-learning type estimators in finite samples, due to the use of importance sampling in constructing the pseudo outcomes. This reflects a bias-variance trade-off. The proposed A-learning method might suffer from a larger variance whereas Q-learning type methods might suffer from a larger bias. This observation is consistent with the findings in the literature on learning DTRs \citep[see e.g.,][]{schulte2014q}. 

%\section{Numerical experiments}
\section{Simulations}\label{sec:syndata}
%\vspace{-0.3cm}
%\subsection{Implementation details}
We evaluate the performance of our method using two synthetic datasets generated by the Open AI Gym environment (see \url{https://gym.openai.com/}) in this section. %Section \ref{sec:syndata}, and a real dataset from a mobile health study in Section \ref{sec:mobilehealth}. 
We consider the following Q-learning type baseline methods: (a) DQN; (b) double DQN (DDQN); (c) quantile DQN (QR-DQN). See Examples \ref{exam:DQN}-\ref{exam:QDQN} for detailed discussion of these algorithms.  
%; (d) BCQ; (e) REM and (f) BEAR. %The programming code is available at \url{https://anonymous.4open.science/r/36d521f7-e500-4956-a318-31e1cb5c389c/}.
As we have commented in Section \ref{sec:PEAL}, our policy optimization procedure at Step 2 is generally applicable to any Q-learning type algorithms. To validate this claim, for each of these Q-learning type methods in (a)-(c), \change{we couple it with sample splitting to compute the initial Q-estimator in Step 2 based on each half of the data}, and apply our proposal in Steps 3-5 to learn an optimal policy. This yields three estimated optimal policies. We denote them by (d) SEAL-DQN, (e) SEAL-DDQN and (f) SEAL-QR-DQN, respectively. \change{Then we contrast them with the corresponding Q-learning type algorithms in (a)-(c) fitted based on the entire offline data}. In addition to these baseline methods, we also consider three recently developed offline policy optimization methods in the computer science literature, including (g) batch-constrained
deep Q-learning \citep[BCQ,][]{fujimoto2019off}, (h) random ensemble mixture  \citep[REM,][]{agarwal2020optimistic} and (i) bootstrapping error accumulation reduction \citep[BEAR,][]{kumar2019stabilizing}. We compare them with the proposed procedure based on QR-DQN, which yields the best performance among (d)-(f). %\change{For fair comparison, we fix the }
%For the three recently developed offline policy optimization methods (d)-(f), we compare them with (I), the proposed procedure based on QR-DQN. 
%All the experiments are run on a single computer instance with 40 Intel(R) Xeon(R) 2.20GHz CPUs. %Additional implementation details are given in Appendix \ref{secaddimp}. %Source code is available at \url{https://github.com/strider2king/SupRL}. 
%Then we contrast this policy with that derived based on the initial Q-estimator. 
%; (d) the proposed method with the initial Q-estimator computed by DQN; 

\begin{figure*}[!t]
	\centering
	\includegraphics[width=\textwidth,height=0.16\textheight]{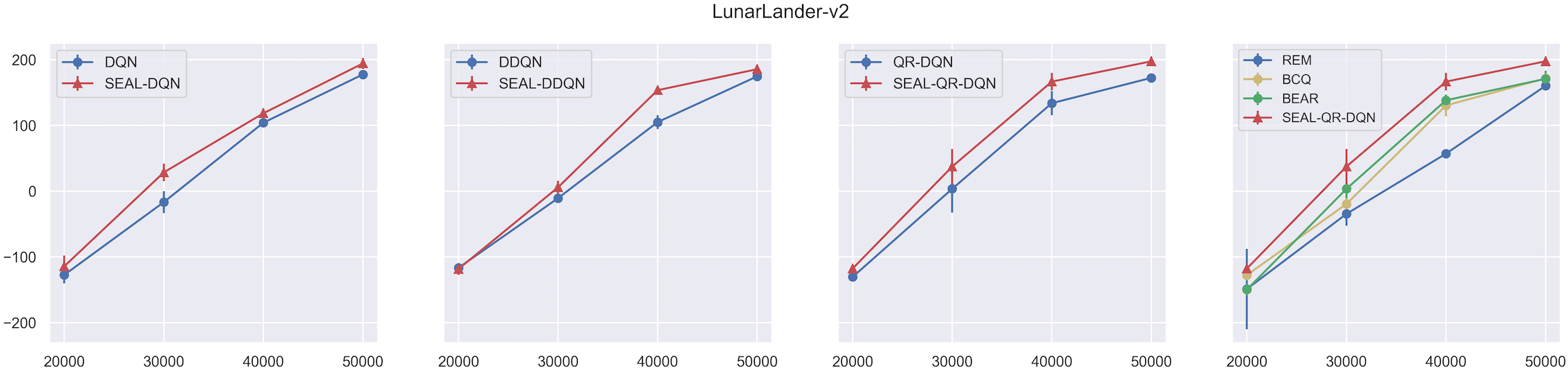}
	%	\vspace{-0.1cm}
	\caption{Synthetic data analysis results for LunarLander-v2. Horizontal axis represents the number of training steps \change{used to train the initial Q-estimator based on half the data as well as the baseline method based on the entire dataset}. Vertical axis represents the average reward of 100 evaluations. %at the last training checkpoint of each experiment. 
	%The data correspond to the first 200 trajectories in the top panels, and the randomly sampled 200 trajectories in the bottom panels. 
	The error bar corresponds to the 95\% confidence interval for the value, constructed based on 10 replications. 
		%The two rows correspond to "First" and "Random" sampling trajectories, respectively. 
	The first three panels compare one baseline Q-learning algorithm (DQN, DDQN, QR-DQN) with our method that uses such a baseline to compute the initial Q-estimator. The last panel compares the proposed algorithm based on QR-DQN against REM, BCQ and BEAR.}
	\label{fig:lunarlander-v2_final}
\end{figure*}

%\vspace{-0.3cm}
%\subsection{Simulations}\label{sec:syndata}
%In this section, 

\subsection{LunarLander-v2}\label{sec:lunar}
We conduct experiments in an OpenAI Gym environment, LunarLander-v2. Detailed description about this environment can be found at \href{https://gym.openai.com/envs/LunarLander-v2/}{LunarLander-v2}. To generate the data, we train a QR-DQN agent $500$K time steps, with learning rate $0.0005$. %(see Appendix \ref{secaddimp} for specification of other tuning parameters). %and the DQN network has two hidden layers with $256$ units each, and $51$ atoms to represent quantiles.  
\change{The estimated policy after $500$K time steps is near optimal and solves the environment (e.g., achieves a score of 200 on average). The state-of-the-art optimal average reward is over 250\footnote{ \url{https://medium.datadriveninvestor.com/training-the-lunar-lander-agent-with-deep-q-learning-and-its-variants-2f7ba63e822c}}. We then terminate the training process, 
%We then 
store all the generated trajectories encountered during the online training process and use them as the offline data. The behavior policy corresponds to the $\epsilon$-greedy policy used to train the online QR-DQN agent with $\epsilon=0.1$. The offline dataset consists of 1089 trajectories.} Each trajectory lasts for 459 time steps on average. The average immediate reward equals 118. 
%and lengths of all the trajectories are $118(154)$ and $459(289)$ with corresponding standard deviation in brackets. 

%We perform two ablation analyses where
%where %we train three base Q networks (DQN, Double DQN(DDQN) and Quantile DQN) and corresponding SupRL on the first 
The training data consist of 
%we train the baseline models in (a)-(c) and our methods (d)-(f) on the datasets that consist of (i) the first 200 trajectories; (ii) 
200 trajectories randomly sampled out of the 1089 trajectories. For each of the estimated optimal policy learned based on (a)-(i), we evaluate its value by computing the mean reward of 100 trajectories generated in the environment under this policy. 
%generating 100 trajectories in the environment under this policy and 
We repeat the entire data generating process, the training and evaluation procedures 10 times with different random seeds. We also vary the number of training steps for the initial Q-estimator and apply the proposed method to each of the estimated Q-functions. \change{For fair comparison, we use the same number of training steps (i.e., 20K, 30K, 40K or 50K) to train the baseline policy.}
%DQN, DDQN and QR-DQN. 
%, and report the mean reward of 100 trajectories generated in the environment via the learned agents' $\epsilon$-greedy policy with $\epsilon = 0.001$. %Detailed hyper parameters are given in the supplementary article. 

Reported in Figure \ref{fig:lunarlander-v2_final} %and \ref{fig:lunarlander-v2_ckpts} 
are the values of the estimated policies computed by (a)-(i) as well as the associated 95\% confidence intervals, with different %combinations of random seed and the data generating process. In Figure \ref{fig:lunarlander-v2_K=2_ckpts} (see Appendix \ref{secaddtablefig}), we depict the values of these policies with different 
number of training steps. %for the initial Q-estimator. 
We summarize our findings as follows: (1) The proposed procedure achieves a larger value compared to the baseline methods in most cases;  %As commented in the introduction, this is due to that many of these baseline methods are derived based on Q-learning. They suffer from a slower rate of convergence, compared to our approach. 
(2) Our improvement is significant in many cases, as suggested by the error bar; (3) All the methods get improved as the number of training steps increases.

\begin{figure*}[!t]
	\centering
	\includegraphics[width=\textwidth,height=0.16\textheight]{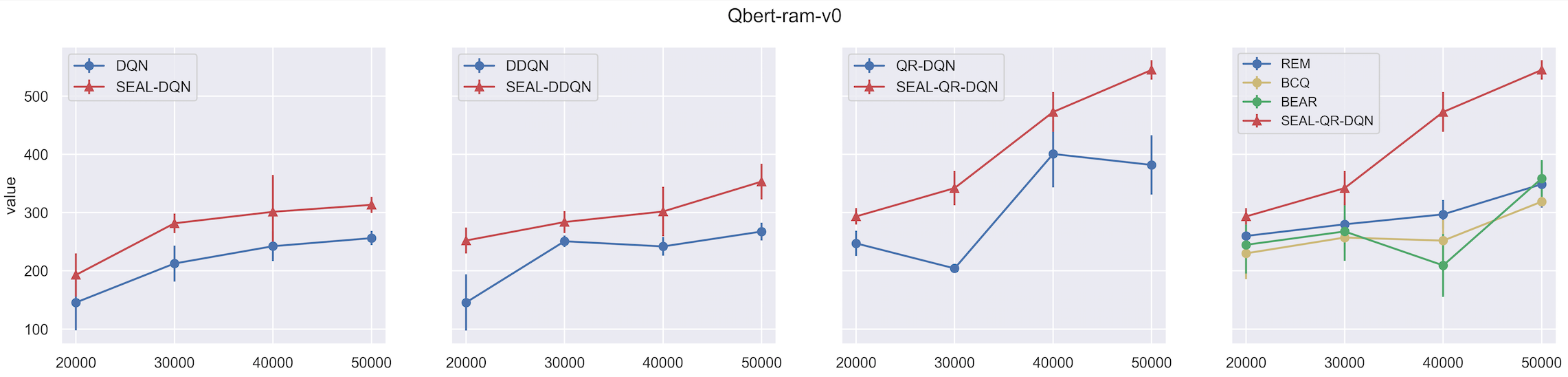}
	%	\vspace{-0.1cm}
	\caption{Synthetic data analysis results for Qbert-ram-v0. %Horizontal axis represents the number of training steps. Vertical axis represents the average reward of 100 evaluations. The error bar corresponds to the 95\% confidence interval for the value, constructed based on 10 replications. 
	Same legend as Figure \ref{fig:lunarlander-v2_final}.} %at the last training checkpoint of each experiment. 
	%The data correspond to the first 200 trajectories in the top panels, and the randomly sampled 200 trajectories in the bottom panels. 
	%The error bar corresponds to the 95\% confidence interval for the value, constructed based on 10 replications. 
	%The two rows correspond to "First" and "Random" sampling trajectories, respectively. 
	%The first three panels compare one baseline Q-learning algorithm (DQN, DDQN, QR-DQN) with our method that uses such a baseline to compute the initial Q-estimator. The last panel compares the proposed algorithm based on QR-DQN against REM, BCQ and BEAR.}
	\label{fig:qbert}
\end{figure*}
%\vspace{-0.1cm}
\begin{figure*}[!t]
	\centering
	\includegraphics[width=\textwidth,height=0.15\textheight]{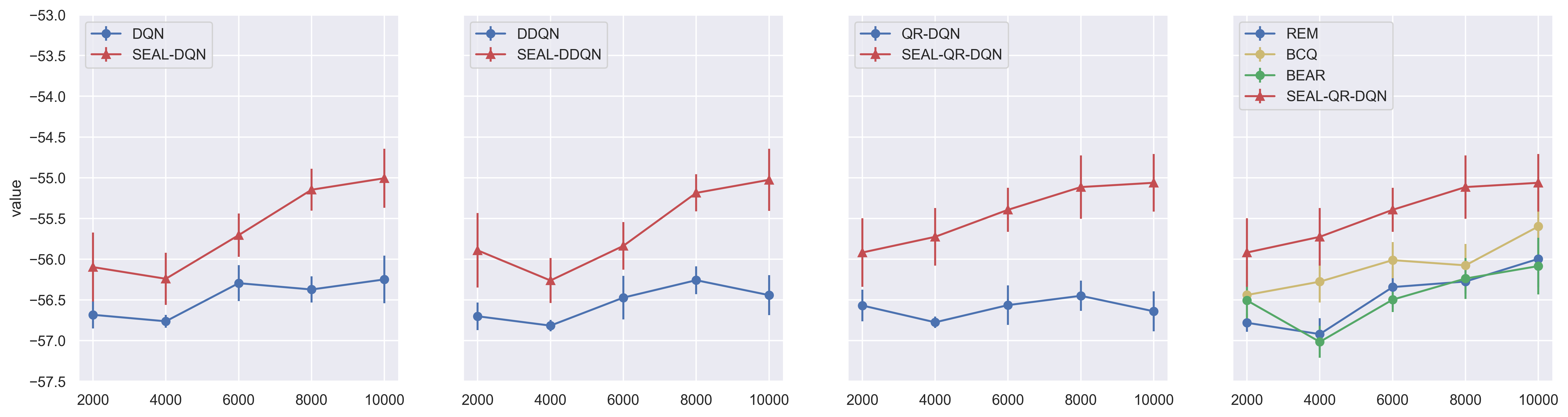}
	%	\vspace{-0.1cm}
	\caption{Real data analysis results. Horizontal axis represents the number of training steps. %Vertical axis represents the average reward of 100 evaluations. %at the last training checkpoint of each experiment. 
		%The data correspond to the first 200 trajectories in the top panels, and the randomly sampled 200 trajectories in the bottom panels. The error bar corresponds to the 95\% confidence interval for the value, constructed based on 10 replications. 
		Same legend as Figure \ref{fig:lunarlander-v2_final}.}
	%The two rows correspond to "First" and "Random" sampling trajectories, respectively. 
	%The first three panels compare one baseline Q-learning algorithm (DQN, DDQN, QR-DQN) with our method that uses such a baseline to compute the initial Q-estimator. The last panel compares the proposed algorithm based on QR-DQN against REM, BCQ and BEAR.}
	\label{fig:realdata}
\end{figure*}
\subsection{Qbert-ram-v0}
We next conduct experiments in another environment, \href{https://gym.openai.com/envs/Qbert-ram-v0/}{Qbert-ram-v0}. The best 100-episode average reward for Qbert-ram-v0 is $586.00 \pm 12.16$. We similarly train a Quantile DQN agent to generate 1373 trajectories. Each trajectory lasts for 364 time steps on average. The average return per trajectory equals 278. We similarly compare our procedures (d)-(f) with the baseline methods (a)-(c) and (g)-(i). Results are depicted in Figure \ref{fig:qbert}. Overall, findings are very similar to those in Section \ref{sec:lunar}. \change{We notice that the performances of some deep Q-learning methods drop when the number of training step increases and cannot even improve after a few more iterations. We discuss this in detail in Section \ref{subsec:pessimistic} to save space. 
%(ii) The proposed method in general achieves better performance compared to recently developed offline learning methods (d)-(f). 
%the proposed methods are more stable and robust in general. For instance, the value of policies estimated via DQN in the top left panel ranges from -100 to 200, across 5 runs. In contrast, values of policies estimated via SupRL-DQN are larger than 100 in all cases. (iii) DDQN and QR-DQN achieve better performance than DQN in general. 
%different combinations of random seed and the number of training steps for DQN, DDQN and QR-DQN. 
%five different random seeds. In Figure \ref{fig:lunarlander-v2_ckpts}, we report

%The results are shown in Figures  - . From the aspect of bootstrapping error incurred by out of distribution states and actions \citep{kumar2019stabilizing}, base Q networks trained on first $200$ trajectories may have larger error than those trained on randomly sampled trajectories, leading to worse performance. As to different base Q networks, DDQN achieves more stable performance than DQN, and Quantile DQN performs even better due to its awareness of uncertainty. Furthermore, SupRL with different base Q networks have robust and stable performances in all cases, which shows SupRL's ability to reduce the bias of Q-estimators and improve performance by its doubly-robustness. %Figure \ref{fig:lunarlander-v2_ckpts} shows that SupRL also performs better during training.

Finally, it can very computationally expensive to implement deep RL algorithms in LunarLander-v2 and Qbert-ram-v0. For instance, in our implementation, it took a few hours to run one simulation. As such, our simulation results are aggregated over 10 runs only. We also remark that beginning with DQN, 5 or less runs are common in the existing RL literature,  as it is often computationally prohibitive to evaluate more runs \citep{agarwal2021deep}; see also the numerical studies in  \citet{mnih2015human,silver2016mastering,kumar2019stabilizing,agarwal2020optimistic}. 
}

%\subsection{Real data examples}
%In this section, we apply our algorithms on two real-world datasets. In both datasets, we use five fold cross-validation to evaluate our method. For each training and testing combination, we train all related agents up to $5$k steps, and apply FQE algorithms on the testing dataset to evaluate values of all agents, every $1$k training steps. Finally, we aggregated these values over different combinations.  We choose $\mathbb{L} = 5$. The above cross-validation procedure is repeated 10 times with different random seeds. We report the average values at different checkpoints and different FQE iterations.
%\vspace{-0.3cm}
\section{The OhioT1DM dataset}
\label{sec:mobilehealth}
%There has been increasing interest in applying RL algorithms to mobile health (mHealth) applications. In this section,
In this section, we use the OhioT1DM Ddataset \citep{marling2018ohiot1dm} to illustrate the usefulness of our new method in moblie health applications. The data contains continuous measurements for six patients with type 1 diabetes over eight weeks. The objective is to learn an optimal policy that maps patients' time-varying covariates into the amount of insulin injected at each time to maximize patients' health status. 

In our experiment, we divide each day of follow-up into one hour intervals and a treatment decision is made every
hour. We consider three important time-varying state variables, including the average blood glucose level
$G_{t}$ during the one hour interval $(t-1, t]$, the carbohydrate estimate for the meal $C_{t}$ during $(t-1, t]$ and $\text{Ex}_{t}$ which measures exercise intensity during $(t-1, t]$. At time $t$, we define the action $A_t$ by discretizing the amount of insulin $\text{In}_t$ injected. The reward $R_t$ is chosen according to the Index of Glycemic Control \citep{rodbard2009interpretation} that is a deterministic function
$G_{t+1}$. Detailed definitions of $A_t$ and $R_t$ are given in Appendix \ref{sec:weight}. 
We will receive a low reward if the patient's average blood glucoses level is outside the range $[80,140]$. Let $X_t=(G_t,C_t,\hbox{Ex}_t)$. We define the state $S_t$ by concatenating measurements over the last four decision points, i.e., $S_t = (X_{t-3}^T, A_{t-3}, \cdots, X_t)^\top$. This ensures the Markov assumption is satisfied \citep{shi2020does}. The number of decision points for each patient in the OhioT1DM dataset ranges from 1119 to 1288. Transitions across different days are treated as independent trajectories.
%We treat transitions within a day as a single trajectory in this study 
This yields 279 trajectories in total.

We use cross-validation to evaluate the performance of different algorithms. Specifically, we apply each of the method in (a)-(i) to the training dataset to learn an optimal policy. Then we apply the fitted Q-evaluation \citep[FQE,][]{le2019batch} algorithm to the testing dataset to evaluate the values of these policies. %(see Appendix \ref{secaddimp} for details). 
FQE is very similar to FQI. It iteratively update the state-action value using supervised learning algorithms. See Algorithm \ref{FQE} in Appendix \ref{sec:weight} for details. In our implementation, we set the function approximator $\mathcal{F}$ to a class of DNN and apply deep learning to update the value. These estimated values are further aggregated over different training/testing combinations. Finally, we repeat this procedure 10 times with different random seeds to further aggregated the values. Results are reported in Figure \ref{fig:realdata}. It can be seen that the proposed method performs significant better than other baseline methods in most cases.

\section{Discussion}\label{sec:moredis}
\change{This section is organized as follows. In Sections \ref{sec:actdis} and \ref{sec:kernel}, we discuss extensions of our proposal to the continuous action space. In Section \ref{subsec:psmooth}, we discuss the $p$-smoothness assumption. Finally, in Section \ref{subsec:pessimistic}, we discuss some offline RL algorithms and the pessimistic principle. 
	%We focus on discrete action spaces throughout the paper. In this section, we outline some methods to handle continuous action space. One possible approach is to first discretize the action space and then apply the proposed method for policy learning. Suppose the action space is one-dimensional, as in personalized dose finding and dynamic pricing. Then we can extend the jump interval-learning method \citep{cai2021jump} to sequential decision making for adaptively discretizing the action space. Another potential approach is to adopt kernel-based methods that leverage treatment proximity to allow continuous actions. In that case, the baseline action can be similarly determined.
	\subsection{Action discretization}\label{sec:actdis}
	One possible approach to handle continuous action space is to first discretize the action space and then apply the proposed method for policy learning. Suppose the action space is one-dimensional, as in personalized dose finding and dynamic pricing. Then we can extend the jump interval-learning method \citep{cai2021jump} to sequential decision making for adaptively discretizing the action space. The baseline action can be similarly determined after we obtain the discretization. 
	
	Specifically, the jump interval learning method was originally designed in contextual bandit settings for deriving interval-valued policies. The main idea is to partition the action space into a set of disjoint intervals such that within each interval, the Q-function is a constant function of the action. In this section, we outline an extension of this method to sequential decision making. Without loss of generally, assume the action space $\mathbb{A}=[0,1]$. To illustrate the method, we assume the transition function is a piecewise function of the action, i.e., there exist a set of disjoint intervals $\mathbb{D}$ that cover $\mathbb{A}$ and satisfy that
	\begin{eqnarray*}
		q(s';a,s)=\sum_{\mathcal{I} \in \mathbb{D}} \mathbb{I}(a\in \mathcal{I}) q_{\mathcal{I}}(s';s),
	\end{eqnarray*}
	for any $s,s'$ and a set of transition functions $q_{\mathcal{I}}$. Under such an assumption, any Q-function is a piecewise function of the action. To identify $\mathbb{D}$, we can couple fitted Q-iteration with jump-interval learning and iteratively solve the following optimization problem based on dynamic programming \citep{friedrich2008complexity},
	\begin{eqnarray*}
		(\widehat{\mathbb{D}}, \{\widehat{Q}_{\mathcal{I}}^{(k)}: \mathcal{I}\in \mathbb{D} \})=\argmin \sum_{\mathcal{I}}\sum_{i,t} \mathbb{I}(A_{i,t}\in \mathcal{I})[R_{i,t}+\gamma \max_{\mathcal{I}} \widehat{Q}_{\mathcal{I}}^{(k-1)}(S_{i,t+1}) -\widehat{Q}_{\mathcal{I}}^{(k)}(S_{i,t}) ]^2 + \gamma |\widehat{\mathbb{D}}|,
	\end{eqnarray*}
	for $k=1,\cdots,K$ and some tuning parameter $\gamma>0$, where $|\widehat{\mathbb{D}}|$ denotes the number of intervals in $\widehat{\mathbb{D}}$. The final output $\widehat{\mathbb{D}}$ yields a set of action intervals, based on which we can define a categorical action variable $A_{i,t}^*$ whose value depends on the interval that the original action $A_{i,t}$ belongs to. Finally, we apply the proposed method to the transformed dataset $\{(S_{i,t}, A_{i,t}^*, R_{i,t})\}_{i,t}$ for policy learning.
	\subsection{Kernel-based method}\label{sec:kernel}
	Another potential approach is to adopt kernel-based methods that leverage treatment proximity to allow continuous actions. In that case, we can apply kernel density estimation to learn the marginal probability density function of the action and set the baseline action to the one that maximizes the estimated density function. 
	
	Specifically, kernel-based methods are commonly used for policy learning and evaluation with continuous action space \citep[see e.g.,][]{chen2016personalized,kallus2018policy,colangelo2020double}. Notice that the proposed method requires to weight each observation by the importance sampling ratio $\mathbb{I}(A_{i,t}=a)/\prob(A_{i,t}=a|S_{i,t})$ to construct the pseudo outcome. When the actions are continuous, the indicator function $\mathbb{I}(A_{i,t}=a)$ equals zero
	almost surely for any $i$ and $t$. Consequently, naively applying the importance sampling ratio would yield a biased estimator. To address this concern, we can employ kernel-based methods to replace the indicator function $\mathbb{I}(A_{i,t}=a)$ with a kernel function $K\{(A-a)/h\}$ for some bandwidth parameter $h$, and probability mass function $\prob(A_{i,t}=a|S_{i,t})$ with the corresponding probability density function. The baseline action can be set to the argmax of $\sum_{i,t} K\{(A_{i,t}-a)/h\}$. 
	
	\subsection{The $p$-smoothness assumption}\label{subsec:psmooth}
	We discuss the $p$-smoothness assumption in this section. First, as commented in the main text, the $p$-smoothness assumption is likely to hold in mobile health studies. In these applications, the state corresponds to some time-varying variables that measure the health status of a given subject. It is expected that the system dynamics would vary smoothly as a function of these  variables. In the literature, a number of papers also use a smooth transition function to simulate the environment in healthcare applications \citep{ertefaie2014,luckett2019,shi2020does,li2022reinforcement}. In addition, the reward is usually a deterministic function of the current state-action pair and future state; see e.g., the definition of the reward in our application in Appendix \ref{sec:weight}. As such, as long as the transition function is $p$-smooth, so is the reward function as well. 
	
	Second, the $p$-smoothness assumption is not likely to hold in many OpenAI gym environments where the state variables are all discrete or the state transition is deterministic. Nonetheless, we remark that it can be seen from our simulation results that the proposed method works better than or comparable to existing Q-learning algorithms. 
	\subsection{Offline RL and the pessimistic principle}\label{subsec:pessimistic}
	As pointed out by one of the referee, it can be seen from Figure \ref{fig:qbert} that the performances of some deep Q-learning methods drop when the number of training step increases and cannot even improve after a few more iterations. We suspect this is due to that some state-action pairs are less explored in the offline dataset, leading to the violation of the positivity assumption (A5). In that case, increasing the number of training steps might overfit the data and worsen the performance of the resulting policy. Similar phenomena have been observed in the existing RL literature \citep[see e.g.,][]{kumar2019stabilizing}. 
	
	Many offline RL algorithms such as BCQ, REM and BEAR adopted the pessimistic principle \citep[see e.g.,][]{levine2020offline,jin2021pessimism,xie2021bellman} and proposed to learn a ``conservative" Q-function by penalizing the values evaluated at those ``out-of-distribution" state-action pairs. These algorithms are shown to outperform the standard Q-learning algorithm in offline domains when (A5) is violated to some extent. We suspect that reducing the number of training steps is equivalent to penalize the Q-function to some extent to avoid overfitting. As such, increasing the number of training steps would make things worse. It is interesting to investigate how to couple the pessimistic principle with our proposal to further improve the estimated policy. However, this is beyond the scope of the current paper. We leave it for future research. }

\bibliographystyle{chicago}
\bibliography{PEAL}

\appendix
\section{Proofs}\label{sec:proof}
We prove Theorems \ref{thm:2} and \ref{thm:3} in this section. Theorem \ref{thm:4} can be proven in a similar manner as Theorem 4 of \cite{shi2020value}. We omit its proof for brevity. 
Throughout this section, we use $c$ and $C$ to denote some generic constants whose values are allowed to vary from place to place. For any two positive sequences $\{a_t\}_{t\ge 1}$ and $\{b_t\}_{t\ge 1}$, we write $a_t\preceq b_t$ if there exists some constant $c>0$ such that $a_t\le cb_t$ for any $t$. The notation $a_t\preceq 1$ means $a_t=O(1)$. %Under the exponential $\beta$-mixing condition, the stochastic process $\{S_t\}_{t\ge 0}$ is stationary. Consequently, the distribution of $S_0$ is the same the limiting distribution fo the process. 

\subsection{Proof of Theorem \ref{thm:2}}
Let $\{\beta(q)\}_{q\ge 0}$ denote the $\beta$-mixing coefficients \citep[see e.g.,][for detailed definitions]{Bradley2005} of the process $\{(S_{t},A_{t},R_{t})\}_{t\ge 0}$. Under the given conditions, we obtain $\beta(q)\to 0$ as $q\to \infty$. Let $b(a|S_t)$ denote the propensity score $\prob(A_t=a|S_t)$ for any $a\in\mathbb{A}$. Let $\widehat{V}^{(\ell)}(s)$ denote the estimated optimal value function $\sum_{a\in \mathbb{A}}\pi^{\widehat{\ell}}(a|s)\widehat{Q}^{(\ell)}(a,s)$ for any $s\in \mathbb{S}$. By definition,  
\begin{eqnarray*}
	\widetilde{Q}_{i,t,a}=\widehat{Q}^{(\ell)}(a,S_{i,t})+\frac{\mathbb{I}(A_{i,t}=a)}{b(a|S_{i,t})}\{R_{i,t}+\gamma \widehat{V}^{(\ell)}(S_{i,t+1})-\widehat{Q}^{(\ell)}(A_{i,t},S_{i,t})\}+\frac{\gamma}{1-\gamma} \widetilde{\eta}_{i,t,a}.
\end{eqnarray*}
Note that
\begin{eqnarray*}
	\Mean [\widetilde{\eta}_{i,t,a}|S_{i,t},\{O_{j,t}\}_{1\le j\le N,0\le t<T}]=\Mean [\eta_{i,t,a}|S_{i,t},\{O_{j,t}\}_{1\le j\le N,0\le t<T}].
\end{eqnarray*}
As such, the conditional mean of $\widetilde{Q}_{i,t,a}$ is the same as $Q_{i,t,a}$, defined as
\begin{eqnarray*}
	Q_{i,t,a}=\widehat{Q}^{(\ell)}(a,S_{i,t})+\frac{\mathbb{I}(A_{i,t}=a)}{b(a|S_{i,t})}\{R_{i,t}+\gamma \widehat{V}^{(\ell)}(S_{i,t+1})-\widehat{Q}^{(\ell)}(A_{i,t},S_{i,t})\}+\frac{\gamma}{1-\gamma} \eta_{i,t,a}.
\end{eqnarray*}
%We decompose $\eta_{i,t,a}$ into $\eta_{i,t,a}^{(1)}+\eta_{i,t,a}^{(2)}+\eta_{i,t,a}^{(3)}$ where
In the following, we break the proof into three steps. In the first step, we show
\begin{eqnarray}\label{eqn:proofthm1eq1}
	\begin{split}
		\Mean \left[\left.\widehat{Q}^{(\ell)}(a,S_{i,t})+\frac{\mathbb{I}(A_{i,t}=a)}{b(a|S_{i,t})}\{R_{i,t}+\gamma\widehat{V}^{(\ell)}(S_{i,t+1}) -\widehat{Q}^{(\ell)}(A_{i,t},S_{i,t})\}\right|S_{i,t},\{O_{j,t}\}_{j\in \mathcal{I}_{\ell}^c,0\le t<T}\right]\\
		=r(a,S_{i,t})+\Mean \{\widehat{V}^{(\ell)}(S_{i,t+1})|S_{i,t},A_{i,t}=a\}=r(a,S_{i,t})+\gamma \int_{s'} \widehat{V}^{(\ell)}(s')q(s';S_{i,t},a)ds'.
	\end{split}	
\end{eqnarray}
Note that $\eta_{i,t,a}$ can be decompose into 
\begin{eqnarray}\label{eqn:etaita}
	\eta_{i,t,a}=\frac{(|\mathcal{I}_{\ell}|-1) T}{|\mathcal{I}_{\ell}| T-1}\eta_{i,t,a,1}+\frac{T-1}{|\mathcal{I}_{\ell}| T-1}\eta_{i,t,a,2},
\end{eqnarray}
where $\eta_{i,t,a,1}$ and $\eta_{i,t,a,2}$ are defined by
\begin{eqnarray*}
	\frac{1}{(|\mathcal{I}_{\ell}|-1) T} \sum_{\substack{i'\in \mathcal{I}_{\ell},i'\neq i\\ 0\le t'<T}} \widehat{\omega}^{(\ell)}(A_{i',t'},S_{i',t'}|a,S_{i,t})\{R_{i',t'}+\gamma \widehat{V}^{(\ell)}(S_{i',t'+1})-\widehat{Q}^{(\ell)}(S_{i',t'},A_{i',t'})\},\\
	\frac{1}{T-1} \sum_{\substack{0\le t'<T\\ t'\neq t }} \widehat{\omega}^{(\ell)}(S_{i,t'};S_{i,t},a)\{R_{i',t'}+\gamma \widehat{V}^{(\ell)}(S_{i,t'+1})-\widehat{Q}^{(\ell)}(S_{i,t'},A_{i,t'})\},
\end{eqnarray*}
respectively.

In the second step, we  show that
\begin{eqnarray}\label{eqn:proofthm1eq2}
	\begin{split}
		\quad\frac{1}{NT}\sum_{\substack{\ell\in \{1,\cdots,\mathbb{L}\}\\i\in \mathcal{I}_{\ell}}}\sum_{0\le t<T}\Mean \left|\Mean \left[\eta_{i,t,a,1}-(1-\gamma)\int_{s'} \{V^{\widehat{\pi}^{(\ell)}}(s')-\widehat{V}^{(\ell)}(s')\}q(s';S_{i,t},a)ds'|S_{i,t}\right]\right|\\\le O(1)(NT)^{-\kappa_1},
	\end{split}
\end{eqnarray}
where $\kappa_1=p/(2p+d)+c_0/2$ and satisfies $p/(2p+d)<\kappa_1<1/2$ under the given conditions on $c_0$, and $O(1)$ denotes some positive constant that is independent of $i,t,a,\ell$. Specifically, we will decompose $\eta_{i,t,a,1}$ into three terms and investigate the conditional mean of each of these terms in detail. %Similarly, we can show
%Under the boundedness assumption on $\{R_{t}\}_{t\ge 0}$, $\widehat{Q}^{(\ell)}$ and $\widehat{\omega}^{(\ell)}$, $\eta_{i,t,a,2}$ is bounded as well. When $N\to \infty$, we have $|\mathcal{I}_{\ell}| \to \infty$ and hence the second term on the RHS of \eqref{eqn:etaita} is $o(1)$. It follows that 
%\begin{eqnarray*}
%	\Mean [\eta_{i,t,a}|S_{i,t},\{O_{j,t}\}_{j\in \mathcal{I}_{\ell}^c,0\le t<T} ]=\int_{s'} \{V^{\widehat{\pi}^{(\ell)}}(s')-\widehat{V}^{(\ell)}(s')\}q(s';a,S_{i,t})ds'+o_p(1).
%\end{eqnarray*}
%It remains to consider the case where $T \to \infty$. 
%In the last step, we  %approximate $\eta_{i,t,a,2}$ by sum of i.i.d. random variables. Then, similar to \eqref{eqn:proofthm1eq2}, we can show 

In the last step, we apply Berbee's coupling lemma \citep[see Lemma 4.1 in][]{Dedecker2002} to show that
\begin{eqnarray}\label{eqn:proofthm1eq3}
	\begin{split}
		\Mean \left|\Mean \left[\eta_{i,t,a,2}-(1-\gamma)\int_{s'} \{V^{\widehat{\pi}^{(\ell)}}(s')-\widehat{V}^{(\ell)}(s')\}q(s';S_{i,t},a)ds'|S_{i,t}\right]\right|
		\\\le O(1) T^{-1}\log T.
	\end{split}
\end{eqnarray}
Combining \eqref{eqn:proofthm1eq1}, \eqref{eqn:proofthm1eq2} and \eqref{eqn:proofthm1eq3} yields that 
\begin{eqnarray*}
	\frac{1}{NT}\sum_{\substack{\ell\in \{1,\cdots,\mathbb{L}\}\\i\in \mathcal{I}_{\ell}}}\sum_{0\le t<T}\Mean \left|\Mean \left[\widetilde{Q}_{i,t,a}|S_{i,t})-r(a,S_{i,t})-\gamma \int_{s'}V^{\widehat{\pi}^{(\ell)}}(s')q(s';S_{i,t},a)ds'|S_{i,t}\right]\right|\\\le O(1)(NT)^{-\kappa_1}.
\end{eqnarray*}
To complete the proof, it remains to show
\begin{eqnarray}\label{eqn:proofthm1eq3.5}
	\max_{i,t}\left|\Mean \left[\int_{s'} \{V^{\widehat{\pi}^{(\ell)}}(s')-V^{\tiny{opt}}(s')\}q(s';S_{i,t},a)ds'|S_{i,t}\right]\right|\le O(1)(NT)^{-\kappa_2},
\end{eqnarray}
for some $\kappa_2>p/(2p+d)$ with probability tending to $1$. Using similar arguments in the proof of Theorem 4 of \citet{shi2020value}, one can show \eqref{eqn:proofthm1eq3.5} holds with $\kappa_2=p(2+2\alpha)/\{(d+2p)(2+\alpha)\}>p/(2p+d)$ under the margin-type condition in (A1). The proof is thus completed. Next, we give detailed proofs for each step. 

%We first consider \eqref{eqn:proofthm1eq1}. 
\subsubsection{Step 1} With some calculations, we have
\begin{eqnarray*}
	\Mean \left[\left.\widehat{Q}^{(\ell)}(a,S_{i,t})+\frac{\mathbb{I}(A_{i,t}=a)}{b(a|S_{i,t})}\{R_{i,t}-\widehat{Q}^{(\ell)}(A_{i,t},S_{i,t})\}\right|S_{i,t},\{O_{j,t}\}_{j\in \mathcal{I}_{\ell}^c,0\le t\le T}\right]\\
	=\widehat{Q}^{(\ell)}(a,S_{i,t})-\widehat{Q}^{(\ell)}(a,S_{i,t})+r(a,S_{i,t})=r(a,S_{i,t}),
\end{eqnarray*}
and
\begin{eqnarray*}
	\Mean \left[\left.\frac{\mathbb{I}(A_{i,t}=a)}{b(a|S_{i,t})}\widehat{V}^{(\ell)}(S_{i,t+1}) \right|S_{i,t},\{O_{j,t}\}_{j\in \mathcal{I}_{\ell}^c,0\le t<T} \right]=\Mean \{\widehat{V}^{(\ell)}(S_{i,t+1})|S_{i,t},A_{i,t}=a\}.
\end{eqnarray*}
This yields \eqref{eqn:proofthm1eq1}. The proof of Step 1 is thus completed. 

\subsubsection{Step 2} We decompose $\eta_{i,t,a,1}$ into $\eta_{i,t,a,3}+\eta_{i,t,a,4}+\eta_{i,t,a,5}$ where
\begin{eqnarray*}
	\eta_{i,t,a,3}&=&\frac{1}{(|\mathcal{I}_{\ell}|-1) T} \sum_{\substack{i'\in \mathcal{I}_{\ell},i'\neq i\\ 0\le t'<T }} \omega^{\widehat{\pi}^{(\ell)}}(A_{i',t'},S_{i',t'}|a,S_{i,t})\widehat{\varepsilon}^{(\ell)}_{i',t'},\\
	\eta_{i,t,a,4}&=&\frac{1}{(|\mathcal{I}_{\ell}|-1) T} \sum_{\substack{i'\in \mathcal{I}_{\ell},i'\neq i\\ 0\le t'<T}} \{\widehat{\omega}^{(\ell)}(A_{i',t'},S_{i',t'}|a,S_{i,t})-\omega^{\widehat{\pi}^{(\ell)}}(A_{i',t'},S_{i',t'}|a,S_{i,t})\}\varepsilon_{i',t'},\\
	\eta_{i,t,a,5}&=&\frac{1}{(|\mathcal{I}_{\ell}|-1) T} \sum_{\substack{i'\in \mathcal{I}_{\ell},i'\neq i\\ 0\le t'<T}} \{\widehat{\omega}^{(\ell)}(A_{i',t'},S_{i',t'}|a,S_{i,t})-\omega^{\widehat{\pi}^{(\ell)}}(A_{i',t'},S_{i',t'}|a,S_{i,t})\}(\widehat{\varepsilon}^{(\ell)}_{i',t'}-\varepsilon_{i',t'}),
\end{eqnarray*}
where $\varepsilon_{i,t}$ denotes the Bellman residual $R_{i,t}+\gamma  V^{\widehat{\pi}^{(\ell)}}(S_{i,t+1})-Q^{\widehat{\pi}^{(\ell)}}(A_{i,t},S_{i,t})$ and $\widehat{\varepsilon}_{i,t}^{(\ell)}$ denotes its estimator by plugging in $\widehat{Q}^{(\ell)}$ and $\widehat{V}^{(\ell)}$ for $Q^{\widehat{\pi}^{(\ell)}}$ and $V^{\widehat{\pi}^{(\ell)}}$, respectively. 

Notice that for any $i'\in \mathcal{I}_{\ell}$, $i'\neq i$, $0\le t'<T$, $O_{i',t'}$ is independent of $O_{i,t}$. With some calculations, we have
\begin{eqnarray}\label{eqn:eta3}
	\Mean (\eta_{i,t,a,3}|S_{i,t})=\int_{s'} \Mean \left[\{V^{\widehat{\pi}^{(\ell)}}(s')-\widehat{V}^{(\ell)}(s')\}q(s';a,S_{i,t})ds'|S_{i,t}\right].
\end{eqnarray}
By Bellman equation, %\citep[see Equation (3.20) in][]{Sutton2018}, 
we have $\Mean (\varepsilon_{i,t}|S_{i,t})=0$ and hence 
\begin{eqnarray}\label{eqn:eta4}
	\Mean [\eta_{i,t,a,4}|S_{i,t},\{O_{j,t}\}_{j\in \mathcal{I}_{\ell}^c,0\le t<T}]=0.
\end{eqnarray}
In view of \eqref{eqn:eta3} and \eqref{eqn:eta4}, to complete the proof of this step, it suffices to show
\begin{eqnarray}\label{eqn:eta5}
	\frac{1}{NT}\sum_{i=1}^N\sum_{t=0}^{T-1}\Mean |\Mean (\eta_{i,t,a,5}|S_{i,t})|=O\{(NT)^{-\kappa_1}\}.
\end{eqnarray} 
%for some $\kappa>p/(2p+d)$. 
Under the stationarity assumption in (A4), we have
\begin{eqnarray*}
	\Mean (\eta_{i,t,a,5}|S_{i,t})
	=\Mean \left[\left.\{\widehat{\omega}^{(\ell)}(A_0,S_{0}|a,S_{i,t})-\omega^{\widehat{\pi}^{(\ell)}}(A_0,S_{0}|a,S_{i,t})\}(\widehat{\varepsilon}^{(\ell)}_{0}-\varepsilon_{0})\right|S_{i,t}\right]\\
	%\end{eqnarray*}
	%It follows from the condition $b$ is bounded away from $0$ that
	%\begin{eqnarray*}
	%	|\Mean (\eta_{i,t,a,5}|S_{i,t})	\le c
	\le \Mean \left[\left.|\widehat{\omega}^{(\ell)}(A_0,S_{0}|a,S_{i,t})-\omega^{\widehat{\pi}^{(\ell)}}(A_0,S_{0}|a,S_{i,t})||\widehat{\varepsilon}^{(\ell)}_{0}-\varepsilon_{0}|\right|S_{i,t} \right].
\end{eqnarray*}
%for some constant $c>0$. 
By Cauchy-Schwarz inequality, we obtain
\begin{eqnarray}\label{eqn:eta5I1I2}
	\begin{split}
		\Mean |\Mean (\eta_{i,t,a,5}|S_{i,t})|^2\le \underbrace{\Mean |\widehat{\varepsilon}^{(\ell)}_{0}-\varepsilon_{0}|^2}_{I_1} \\\times  \underbrace{\Mean |\widehat{\omega}^{(\ell)}(A_0,S_{0}|a,S_{i,t})-\omega^{\widehat{\pi}^{(\ell)}}(A_0,S_{0}|a,S_{i,t})|^2 }_{I_2}. 
	\end{split}	
\end{eqnarray}
By definition, we have
\begin{eqnarray*}
	|\widehat{\varepsilon}_{0}^{(\ell)}-\varepsilon_{0}|\le |\widehat{V}^{(\ell)}(S_{1})-V^{\widehat{\pi}^{(\ell)}}(S_{1})|+|\widehat{Q}^{(\ell)}(A_0,S_0)-Q^{\widehat{\pi}^{(\ell)}}(A_{0},S_{0})|.
\end{eqnarray*}
It follows from Cauchy-Schwarz inequality and the stationarity condition that 
\begin{eqnarray*}
	&&I_1\le 2\Mean |\widehat{V}^{(\ell)}(S_{1})-V^{\widehat{\pi}^{(\ell)}}(S_{1})|^2+ 2|\widehat{Q}^{(\ell)}(A_0,S_0)-Q^{\widehat{\pi}^{(\ell)}}(A_{0},S_{0})|^2\\
	&=& 2\Mean |\widehat{V}^{(\ell)}(S_{0})-V^{\widehat{\pi}^{(\ell)}}(S_{0})|^2+ 2\Mean|\widehat{Q}^{(\ell)}(A_0,S_0)-Q^{\widehat{\pi}^{(\ell)}}(A_{0},S_{0})|^2\\&\le& \max_{a} 4\Mean |\widehat{Q}^{(\ell)}(a,S_{0})-Q^{\widehat{\pi}^{(\ell)}}(a,S_{0})|^2.
\end{eqnarray*}
Since $p_{\infty}$ is uniformly bounded away from zero, it follows that
\begin{eqnarray}\label{pinftyboundedaway}
	I_1\le c\Mean_{(a,s)\sim p_{\infty}} |\widehat{Q}^{(\ell)}(a,s)-Q^{\widehat{\pi}^{(\ell)}}(a,s)|^2,
\end{eqnarray}
for some constant $c>0$. Theorem 4 of \cite{shi2020value} showed that the value under an estimated optimal policy computed by Q-learning type estimators converges at a faster rate than the estimated Q-function under (A1). It follows that
\begin{eqnarray*}
	I_1\le 2c\Mean_{(a,s)\sim p_{\infty}} |\widehat{Q}^{(\ell)}(a,s)-Q^{\tiny{opt}}(a,s)|^2.
\end{eqnarray*}
Similar to \eqref{pinftyboundedaway}, we have
\begin{eqnarray*}
	I_2\le C\Mean_{(a,s)\sim p_{\infty},(a',s')\sim p_{\infty}} |\widehat{\omega}^{(\ell)}(a',s'|a,s)-\omega^{\widehat{\pi}^{(\ell)}}(a',s'|a,s)|^2,
\end{eqnarray*}
for some constant $C>0$. 
Combining these together with \eqref{eqn:eta5I1I2} yields
\begin{eqnarray*}
	\Mean |\Mean (\eta_{i,t,a,5}|S_{i,t})|^2\le 2cC \Mean_{(a,s)\sim p_{\infty}} |\widehat{Q}^{(\ell)}(a,s)-Q^{\tiny{opt}}(a,s)|^2 \\\times \Mean_{(a,s)\sim p_{\infty},(a',s')\sim p_{\infty}} |\widehat{\omega}^{(\ell)}(a',s'|a,s)-\omega^{\widehat{\pi}^{(\ell)}}(a',s'|a,s)|^2.
\end{eqnarray*}
Under the given conditions in (A3), we 
%we require either $\max_{a,\ell} \int_{s'} |\widehat{Q}^{(\ell)}(s',a)-Q^{\tiny{opt}}(s',a)|^2\mathbb{F}_0(ds')=o_p(1)$ or $\sup_{s,a,\ell} \int_{s'} |\widehat{\omega}^{(\ell)}(s';s,a)-\omega^{\widehat{\pi}^{(\ell)}}(s';s,a)|^2\mathbb{F}_0(ds')=o_p(1)$. Consequently, we 
obtain \eqref{eqn:eta5}. This completes the proof of this step. 

\subsection{Step 3} For a given integer $q>1$, we decompose $\eta_{i,t,a,2}$ into $\eta_{i,t,a,6}+\eta_{i,t,a,7}$ where
\begin{eqnarray*}
	\eta_{i,t,a,6}=\frac{1}{T-1} \sum_{|t'-t|\le q} \widehat{\omega}^{(\ell)}(A_{i',t'},S_{i,t'}|a,S_{i,t})\widehat{\varepsilon}^{(\ell)}_{i,t'},\\
	\eta_{i,t,a,7}=\frac{1}{T-1} \sum_{|t'-t|> q} \widehat{\omega}^{(\ell)}(A_{i',t'},S_{i,t'}|a,S_{i,t})\widehat{\varepsilon}^{(\ell)}_{i,t'}.
\end{eqnarray*}
%As $T$ diverges, $\eta_{i,t,a,6}$ is $o(1)$ under the given conditions. In the following, we show
%\begin{eqnarray*}
%	\Mean [\eta_{i,t,a,7}|S_{i,t},\{O_{j,t}\}_{j\in \mathcal{I}_{\ell}^c,0\le t<T}]=\int_{s'} \{V^{\widehat{\pi}^{(\ell)}}(s')-\widehat{V}^{(\ell)}(s')\}q(s';a,S_{i,t})ds'+o_p(1)+R(q),
%\end{eqnarray*}
%where $R(q)\to 0$ as $q\to \infty$. The proof is hence completed. 
For any $t'$ such that $|t'-t|>q$, we have $|t'+1-t|\ge q$. Consequently, the $\beta$-mixing coefficient between the two $\sigma$-algebras  $\sigma(S_{i,t})$ and $\sigma(S_{i,t'},A_{i,t'},R_{i,t'},S_{i,t'+1}))$ are upper bounded by $\beta(q)$, under the given conditions. By Berbee's lemma, there exists an identical copy $(\widetilde{S}_{i,t'},\widetilde{A}_{i,t'},\widetilde{R}_{i,t'},\widetilde{S}_{i,t'+1})$ of $(S_{i,t'},A_{i,t'},R_{i,t'},S_{i,t'+1})$ that is independent of $S_{i,t}$ such that
\begin{eqnarray}\label{eqn:Berbee}
	\prob\{(\widetilde{S}_{i,t'},\widetilde{A}_{i,t'},\widetilde{R}_{i,t'},\widetilde{S}_{i,t'+1})\neq (S_{i,t'},A_{i,t'},R_{i,t'},S_{i,t'+1})\}\le \beta(q).
\end{eqnarray} 
Consequently, $\eta_{i,t,a,7}$ can be decomposed into
\begin{eqnarray*}
	\frac{1}{T-1} \sum_{|t'-t|> q} \widehat{\omega}^{(\ell)}(A_{i',t'},S_{i,t'};a,S_{i,t})\widehat{\varepsilon}^{(\ell)}_{i,t'}\mathbb{I}\{(S_{i,t'},A_{i,t'},S_{i,t'+1})=(\widetilde{S}_{i,t'},\widetilde{A}_{i,t'},\widetilde{S}_{i,t'+1})\}\\
	+\frac{1}{T-1} \sum_{|t'-t|> q} \widehat{\omega}^{(\ell)}(A_{i',t'},S_{i,t'};a,S_{i,t})\widehat{\varepsilon}^{(\ell)}_{i,t'}\mathbb{I}\{(S_{i,t'},A_{i,t'},S_{i,t'+1})\neq (\widetilde{S}_{i,t'},\widetilde{A}_{i,t'},\widetilde{S}_{i,t'+1})\},
\end{eqnarray*}
and is equal to
\begin{eqnarray*}	
	=\frac{1}{T-1} \sum_{|t'-t|> q} \widehat{\omega}^{(\ell)}(\widetilde{A}_{i',t'},\widetilde{S}_{i,t'}|a,S_{i,t})\{\widetilde{R}_{i,t}+\gamma \widehat{V}^{(\ell)}(\widetilde{S}_{i,t'+1})-\widehat{Q}^{(\ell)}(\widetilde{S}_{i,t},\widetilde{A}_{i,t})\}\\
	+\frac{1}{T-1} \sum_{|t'-t|> q} \widehat{\omega}^{(\ell)}(\widetilde{A}_{i',t'},\widetilde{S}_{i,t'};a,S_{i,t})\widehat{\varepsilon}^{(\ell)}_{i,t'}\mathbb{I}\{(S_{i,t'},A_{i,t'},S_{i,t'+1})\neq (\widetilde{S}_{i,t'},\widetilde{A}_{i,t'},\widetilde{S}_{i,t'+1})\}\\
	-\frac{1}{T-1} \sum_{|t'-t|> q} \widehat{\omega}^{(\ell)}(\widetilde{A}_{i',t'},\widetilde{S}_{i,t'};a,S_{i,t})\{\widetilde{R}_{i,t}+\gamma \widehat{V}^{(\ell)}(\widetilde{S}_{i,t'+1})-\widehat{Q}^{(\ell)}(\widetilde{S}_{i,t},\widetilde{A}_{i,t})\}\\
	\times \mathbb{I}\{(S_{i,t'},A_{i,t'},S_{i,t'+1})\neq (\widetilde{S}_{i,t'},\widetilde{A}_{i,t'},\widetilde{S}_{i,t'+1})\}.
\end{eqnarray*}
By \eqref{eqn:Berbee} and the boundedness assumption, the conditional expectation of the last two terms are bounded by $c \beta(q)$, for some constant $c>0$. As $\beta$ is exponentially decreasing, by setting $q$ to be proportional $\log T$, the last two terms decay at a rate of $T^{-\kappa}$ for any $\kappa>0$. Using similar arguments in Step 2 of the proof, the conditional expectation of the first term is asymptotically equivalent to $\Mean [\int_{s'} \{V^{\widehat{\pi}^{(\ell)}}(s')-\widehat{V}^{(\ell)}(s')\}q(s';a,S_{i,t})ds'|S_{i,t}]$ and can be similarly bounded.

In addition, by setting $q$ to be proportional to $\log T$, we have $\eta_{i,t,a,6}=O(T^{-1}\log T)$ The proof is hence completed. 

\subsection{Proof of Theorem \ref{thm:3}}\label{sec:proof2}
%\begin{figure}[!t]
%	\centering
%	\includegraphics[width=5cm]{MLP1.png}
%	\caption{Illustration of DNN with two hidden layers, $m_0=2$, $m_1=m_2=3$. Here $u$ is the input, $A^{(\ell)}$ and $b^{(\ell)}$ denote the corresponding parameters to produce the linear transformation for the $(\ell-1)$th layer.}\label{fig:DNN}
%\end{figure}
%We first specify the DNN class $\mathcal{T}$. 
%For a given action $a\neq a_0$, let $\mathcal{T}_a$ be some DNN class of functions $\{\tau(s,a):s\in \mathbb{S}\}$ whose detailed form will be specified later. We define $\mathcal{T}=\{\tau(a,s)=\tau_a(s):\tau_a\in \mathcal{T}_a\}$. We require $\mathcal{T}$ to be a  uniformly bounded function class. Let $\widehat{\tau}_a$ denote the empirical minimizer of $\argmin_{\tau \in \mathcal{T}_a} \sum_{i=1}^N \sum_{t=0}^{T-1} \{\widetilde{\tau}_{i,t,a}-\tau_a(S_{i,t})\}^2$. 
We use a shorthand and write $\widehat{\tau}(a,\bullet)$ as $\widehat{\tau}_a(\bullet)$. Under the stationarity assumption, it suffices to show $\Mean |\widehat{\tau}_a(S_0)-\tau^{\tiny{opt}}(a,S_0)|^2$ converges at a rate of $O\{(NT)^{-2\kappa_0}\}$ for some  constant $\kappa_0$ such that $p/(2p+d)<\kappa_0<1/2$, for any action $a\neq a_0$. 

We begin by specifying the DNN class $\mathcal{T}_a$. We consider using ReLU as the activation function. Let $\theta$ denote a vector of parameters involved in the DNN. We use $W$ to denote the total numbers of parameters in the DNN. Let $L$ denote the number of layers in the DNN.  Without loss of generality, we assume $\sup_{a,s}|\tau^{\tiny{opt}}(a,s)|\le M$ for any constant $M>0$. For any sufficiently small $\epsilon>0$, there exists a DNN function class 
\begin{eqnarray*}
	\mathcal{T}=\left\{\tau_a(\bullet,\theta): W\le \bar{C}\epsilon^{-d/p^*}(\log \epsilon^{-1}+1), L\le \bar{C}(\log \epsilon^{-1}+1), \sup_{a,s}|\tau_a(s,\theta)|\le M \right\},
\end{eqnarray*}
for some constant $\bar{C}>0$ such that there exists some $\tau^*_a$ belonging to $\mathcal{T}_a$ that approximates $\tau^{\tiny{opt}}(a,\bullet)$ with the uniform approximation error upper bounded by $\epsilon$. See e.g., Lemma 7 of \citet{farrell2021deep}. We set $\epsilon=(NT)^{-\kappa_0}$. It suffices to show that $\Mean |\widehat{\tau}_a(S_0)-\tau^{*}_a(S_0)^2|$ converges at a rate of $O\{(NT)^{-\kappa_0}\}$.

We aim to upper bound the probability
\begin{eqnarray}\label{eqn:upperboundprob}
	\prob\left( \Mean \{|\widehat{\tau}_a(S_0)-\tau_a^*(S_0)|^2|\widehat{\tau}_a\}>k_0^2(NT)^{-2\kappa_0} \right),
\end{eqnarray} 
for some positive integer $k_0$. The detailed requirements of $k_0$ and $\kappa_0$ will be specified later. Notice that
\begin{eqnarray}\label{eqn:1}
	\begin{split}
		\Mean \{|\widehat{\tau}_a(S_0)-\tau_a^{*}(S_0)|^2=\Mean \{|\widehat{\tau}_a(S_0)-\tau_a^{*}(S_0)|^2\mathbb{I}\{|\widehat{\tau}_a(S_0)-\tau_a^{*}(S_0)|^2>k_0^2(NT)^{-2\kappa_0}\}\\
		+\Mean \{|\widehat{\tau}_a(S_0)-\tau_a^*(S_0)|^2\mathbb{I}\{|\widehat{\tau}_a(S_0)-\tau_a^*(S_0)|^2\le k_0^2(NT)^{-2\kappa_0}\}.
	\end{split}	
\end{eqnarray} 
Suppose we can show the probability \eqref{eqn:upperboundprob} can be upper bounded by $O(N^{-1}T^{-1})$. The second term on the right-hand-side (RHS) of \eqref{eqn:1} can be upper bounded by $k_0^2(NT)^{-2\kappa_0}$. Under the $p$-smoothness assumption, the optimal Q-function is continuous over the state space. As such, it is uniformly bounded. So is the optimal contrast function. The first term is $O(N^{-1}T^{-1})$ given that the class $\mathcal{T}_a$ is a bounded function class. This yields the desired rate of convergence. The proof is thus completed.

%To bound \eqref{eqn:upperboundprob},  %The detailed form of $\epsilon$ will be specified later. 

%Consider a fixed action $a$. 
It remains to show that the probability \eqref{eqn:upperboundprob} can be upper bounded by $O(N^{-1}T^{-1})$. Notice that \eqref{eqn:upperboundprob} is upper bounded by
\begin{eqnarray}\label{eqn:upperboundprob0}
	\sum_{k\ge k_0} \prob\left( k^2(NT)^{-2\kappa_0}<\Mean \{|\widehat{\tau}_a(S_0)-\tau_a^*(S_0)|^2|\widehat{\tau}_a\}\le (k+1)^2(NT)^{-2\kappa_0} \right),
\end{eqnarray} 
by Bonferroni's inequality. 

Consider an integer $k\ge k_0$. %We focus on bounding the probability
Let $\mathcal{A}_k$ denote the event, 
\begin{eqnarray*}%\label{eqn:upperboundprob1}
	\prob\left( k^2(NT)^{-2\kappa_0}<\Mean \{|\widehat{\tau}_a(S_0)-\tau^{*}_a(S_0)|^2|\widehat{\tau}_a\}\le (k+1)^2(NT)^{-2\kappa_0} \right).
\end{eqnarray*} 
We focus on evaluating the following stochastic error term,
\begin{eqnarray}\label{eqn:var}
	\left|\frac{2}{NT}\sum_{i=1}^N \sum_{t=0}^{T-1} \{\widetilde{\tau}_{i,t,a}-\tau_a^{*}(S_{i,t})\}\{\widehat{\tau}_a(S_{i,t})-\tau_a^*(S_{i,t})\} \right|,
\end{eqnarray}
under the event defined in $\mathcal{A}_k$.

%The arguments are very similar to that of Theorem 1 of \citet{imaizumi2019deep}. We provide a sketch of the proof for brevity. Let $n=N/L$. 
%In the following, we use $O(1)$ to denote some positive constant, whose value is allowed to vary from place to place. 
Since the number of folds $\mathbb{L}$ and the action space is finite, it suffices to bound
\begin{eqnarray}\label{eqn:-1}
	\left|\frac{2}{nT}\sum_{i\in \mathcal{I}_{\ell}}\sum_{t=0}^{T-1} \{\widetilde{\tau}_{i,t,a}-\tau_a^*(S_{i,t})\}\{\widehat{\tau}_a(S_{i,t})-\tau_a^*(S_{i,t})\}\right|.
\end{eqnarray}
The above bound can be upper bounded by the sum of the following three terms,
\begin{eqnarray}\label{eqn:eta1}
	\left|\frac{2}{nT}\sum_{i\in \mathcal{I}_{\ell}}\sum_{t=0}^{T-1} \{\Mean (\widetilde{\tau}_{i,t,a}|S_{i,t})-\tau_a^*(S_{i,t})\}\{\widehat{\tau}_a(S_{i,t})-\tau_a^*(S_{i,t})\}\right|,\\ \label{eqn:eta2}
	\left|\frac{2}{nT}\sum_{i\in \mathcal{I}_{\ell}}\sum_{t=0}^{T-1} (\widetilde{\tau}_{i,t,a}- \tau_{i,t,a})\{\widehat{\tau}_a(S_{i,t})-\tau_a^*(S_{i,t})\}\right|,\\\label{eqn:eta6}
	\left|\frac{2}{nT}\sum_{i\in \mathcal{I}_{\ell}}\sum_{t=0}^{T-1} \{\tau_{i,t,a}-\Mean (\tau_{i,t,a}|S_{i,t})\}\{\widehat{\tau}_a(S_{i,t})-\tau_a^*(S_{i,t})\}\right|,
\end{eqnarray}
where $\tau_{i,t,a}$ is a version of $\widetilde{\tau}_{i,t,a}$ with $\widetilde{\eta}_{i,t,a}$ replaced by $\eta_{i,t,a}$.

By Cauchy-Schwarz inequality, \eqref{eqn:eta1} is upper bounded by
\begin{eqnarray}\label{eqn-1}
	%\begin{split}
	\frac{2}{\varepsilon nT}\sum_{i\in \mathcal{I}_{\ell}}\sum_{t=0}^{T-1}|\Mean (\widetilde{\tau}_{i,t,a}|S_{i,t})-\tau_a^*(S_{i,t})|^2+\frac{2\varepsilon}{nT}\sum_{i\in \mathcal{I}_{\ell}}\sum_{t=0}^{T-1} |\widehat{\tau}_a(S_{i,t})-\tau_a^*(S_{i,t})|^2, %\\\le& \frac{O(1)}{\varepsilon}  (nT)^{-2\kappa}+ \varepsilon\left|\frac{2}{nT}\sum_{i\in \mathcal{I}_{\ell}}\sum_{t=0}^{T-1} |\widehat{\tau}_a(S_{i,t})-\tau_a^*(S_{i,t})|^2 \right|,
	%\end{split}	
\end{eqnarray}
for any sufficiently small $\varepsilon>0$. 
It follows from Theorem \ref{thm:2} and the definition of $\tau_a^*$ that the expectation of the first term of \eqref{eqn-1} is of the order $O\{(nT)^{-2\kappa}\}$ with $\kappa=\min(p/(2p+d)+c_0/2, p(2+2\alpha)/\{(2p+d)(2+\alpha)\})$. Without loss of generality, assume $\kappa<1/2$. Under the given conditions, $\{\widetilde{\tau}_{i,t,a}\}_{i,t,a}$ are uniformly bounded. So are their conditional expectations. In addition, for any $(i_1,t_1)$ and $(i_2,t_2)$, we can similarly show that the absolute value of the covariance
\begin{eqnarray*}
	|\Cov\{ |\Mean (\widetilde{\tau}_{i_1,t_1,a}|S_{i_1,t_1})-\tau_a^*(S_{i_2,t_2})|^2, |\Mean (\widetilde{\tau}_{i_2,t_2,a}|S_{i_2,t_2})-\tau_a^*(S_{i_2,t_2})|^2\}|,
\end{eqnarray*}
is of the order $O\{(nT)^{-4\kappa}\}$ as well. Since $\{S_{i,t}\}_{i,t}$ are exponentially $\beta$-mixing, applying the concentration inequality for sums of $\beta$-mixing random variables \citep[see e.g., Theorem 4.2][]{Chen2015} yields that the first term is upper bounded by $O(1)(nT)^{-2\kappa}$ with probability at least $1-O(N^{-1}T^{-1})$. We use $\mathcal{A}^*$ denote this event. In other words, on the set $\mathcal{A}^*$, \eqref{eqn:eta1} is upper bounded by
\begin{eqnarray}
	\begin{split}
		&\frac{2}{nT}\sum_{i\in \mathcal{I}_{\ell}}\sum_{t=0}^{T-1}|\Mean (\widetilde{\tau}_{i,t,a}|S_{i,t})-\tau_a^*(S_{i,t})|^2+\frac{2}{nT}\sum_{i\in \mathcal{I}_{\ell}}\sum_{t=0}^{T-1} |\widehat{\tau}_a(S_{i,t})-\tau_a^*(S_{i,t})|^2 \\\le& \frac{O(1)}{\varepsilon}  (nT)^{-2\kappa}+ \varepsilon\left|\frac{2}{nT}\sum_{i\in \mathcal{I}_{\ell}}\sum_{t=0}^{T-1} |\widehat{\tau}_a(S_{i,t})-\tau_a^*(S_{i,t})|^2 \right|,
	\end{split}	
\end{eqnarray}
for any sufficiently small constant $\varepsilon>0$. %and some $\kappa>p/(2p+d)$. 

We next consider \eqref{eqn:eta2}. Note that for any $i,t$, the set of variables $\{\widetilde{\tau}_{i,t,a}-\tau_{i,t,a}\}_{i,t}$ are mean-zero and independent conditional on the observed data. We aim to apply the empirical process theory \citep{van1996} to bound \eqref{eqn:eta2}. 
%Specifically, for any $\tau\in \mathcal{T}$, let $\sigma^2(\tau,a)=\Mean |\tau(S_0,a)-\tau^*(S_0,a)|^2$. Under the stationarity assumption, $\Mean |\tau(S_0,a)-\tau^*(S_0,a)|^2= \Mean |\tau(S_{i,t},a)-\tau^*(S_{i,t},a)|^2$ for any $i$ and $t$. We consider a scaled version of \eqref{eqn:eta2}, defined by
%\begin{eqnarray}\label{eqn:eta-1}
%	\frac{2}{\sigma(\widehat{\tau},a)}\left|\frac{1}{nT}\sum_{i\in \mathcal{I}_{\ell}}\sum_{t=0}^{T-1} (\widetilde{\tau}_{i,t,a}- \tau_{i,t,a})\{\widehat{\tau}_a(S_{i,t})-\tau^*(S_{i,t},a)\}\right|.
%\end{eqnarray}
Under the event defined in $\mathcal{A}_k$, we have $\widehat{\tau}_a\in \mathcal{T}_{a,k}$ where
\begin{eqnarray*}
	\mathcal{T}_{a,k}=\{\tau_a: k^2(NT)^{-2\kappa_0}<\Mean |\tau_a(S_0)-\tau^{*}_a(S_0)|^2|\le (k+1)^2(NT)^{-2\kappa_0} \}.
\end{eqnarray*}
As such, \eqref{eqn:eta2} can be upper bounded by
\begin{eqnarray}\label{eqn:eta-1}
	\sup_{\tau_a \in \mathcal{T}_a,k}\mathbb{Z}(\tau,a)=\sup_{\tau_a\in \mathcal{T}_a,k}\left|\frac{2}{nT}\sum_{i\in \mathcal{I}_{\ell}}\sum_{t=0}^{T-1} (\widetilde{\tau}_{i,t,a}- \tau_{i,t,a})\{\tau_a(S_{i,t})-\tau_a^*(S_{i,t})\}\right|.
\end{eqnarray}
We first consider bounding the expectation $\Mean \sup_{\tau_a \in \mathcal{T}_{a,k}}\mathbb{Z}(\tau,a)$. Toward that end, we apply the maximal inequality developed in Corollary 5.1 of \citet{cherno2014}. Under the given conditions, the set of variables $\{\widetilde{\tau}_{i,t,a}\}_{i,t,a}$, $\{\tau_{i,t,a}\}_{i,t,a}$ are uniformly bounded. Let $M^*>0$ be the upper bound. As such, the absolute value of each summand in \eqref{eqn:eta2} is upper bounded by $4(nT)^{-1} MM^*$. Moreover, the class of the sum over $\mathcal{T}$ belongs to the VC subgraph class with VC-index upper bounded by $O(WL\log W)=O\{ (NT)^{\kappa_0d/p^*}\log^3 (NT) \}$. See Lemmas 4-6 in \citet{farrell2021deep}. It follows that
\begin{eqnarray}\label{eqn:supz}
	\begin{split}
		\Mean \sup_{\tau_a\in \mathcal{T}_{a,k}}\mathbb{Z}(\tau,a)\preceq (NT)^{-1/2}\left\{k\sqrt{(NT)^{-2\kappa_0+\kappa_0d/p^*} \log^4(NT)}\right.\\\left.+(NT)^{\kappa_0d/p^*-1/2}\log^4 (NT) \right\}. 
	\end{split}
\end{eqnarray}
%
%\begin{eqnarray*}
%	\delta=\max \left\{S^{1/2}(NT)^{-1/2}, (nT)^{-1/2}\left(\sum_{i\in \mathcal{I}_{\ell}}\sum_{t=0}^{T-1} |\widehat{\tau}_a(S_{i,t})-\tau(S_{i,t},a)|^2\right)^{1/2} \right\}.
%\end{eqnarray*}
%Define the function class 
%\begin{eqnarray*}
%	\mathcal{T}_{\delta}=\left\{\widehat{\tau}-\tau: (nT)^{-1} \sum_{i\in \mathcal{I}_{\ell}} \sum_{t=0}^{T-1} |\widehat{\tau}_a(S_{i,t})-\tau(S_{i,t},a)|^2\le \delta^2,\tau \in \mathcal{T} \right\}.
%\end{eqnarray*}
%Using similar arguments in Section B.2 of \citet{imaizumi2019deep}, we can show
%\begin{eqnarray}\label{eqn:0}
%	\Mean \left[\sup_{f \in \mathcal{T}_{\delta}} 	\left|\frac{2}{nT}\sum_{i\in \mathcal{I}_{\ell}}\sum_{t=0}^{T-1} (\widetilde{\tau}_{i,t,a}- \tau_{i,t,a})f(S_{i,t},a)\right| |\{O_{i,t}\}_{i,t}\right]\le O(1) \frac{\delta \sqrt{S}}{\sqrt{nT}} \left\{1+\log \frac{L}{B\delta}\right\}.
%\end{eqnarray}
Under the given conditions, $\mathcal{T}_{a,k}$ is a bounded function class. It follows from Theorem 2 of \citet{adamczak2008tail} that for any $u>0$, we have with probability $1-\exp(-u)$ that 
\begin{eqnarray*}
	&&\sup_{\tau\in \mathcal{T}_{a,k}} 	\left|\frac{2}{nT}\sum_{i\in \mathcal{I}_{\ell}}\sum_{t=0}^{T-1} (\widetilde{\tau}_{i,t,a}- \tau_{i,t,a})\{ \tau_a(S_{i,t})-\tau_a^*(S_{i,t}) \}\right|\\
	&\le& 	2\Mean \left[\sup_{\tau\in \mathcal{T}_{a,k}} 	\left|\frac{2}{nT}\sum_{i\in \mathcal{I}_{\ell}}\sum_{t=0}^{T-1} (\widetilde{\tau}_{i,t,a}- \tau_{i,t,a})\{ \tau_a(S_{i,t})-\tau_a^*(S_{i,t}) \} \right| |\{O_{i,t}\}_{(i,t) \in \mathcal{I}_{\ell}^c }\right]\\&+& O(1)  (NT)^{-1/2} \{k\sqrt{u}(NT)^{-\kappa_0}+(NT)^{-1/2}u\}.
\end{eqnarray*}
This together with \eqref{eqn:supz} yields 
\begin{eqnarray*}
	\sup_{\tau\in \mathcal{T}_{a,k}} 	\left|\frac{2}{nT}\sum_{i\in \mathcal{I}_{\ell}}\sum_{t=0}^{T-1} (\widetilde{\tau}_{i,t,a}- \tau_{i,t,a})\{\tau_a(S_{i,t})-\tau_a^*(S_{i,t})\}\right|\le O(1) (NT)^{-1/2}\{k\sqrt{u}(NT)^{-\kappa_0}\\+(NT)^{-1/2}u+
	k\sqrt{(NT)^{\kappa_0d/p^*-2\kappa_0} \log^4(NT)}+(NT)^{\kappa_0d/p^*-1/2}\log^4 (NT) \},
\end{eqnarray*}
with probability at least $1-\exp(-u)$. Set $u=2\log (kNT)$, \eqref{eqn:eta2} is upper bounded by
\begin{eqnarray}\label{eqn:eta0}
	\begin{split}
		\frac{O(1)}{\sqrt{NT}}\left\{k\sqrt{(NT)^{\kappa_0d/p^*-2\kappa_0} \log^4(NT)}+(NT)^{\kappa_0d/p^*-1/2}\log^4 (NT)\right.\\
		\left.+k(NT)^{-\kappa_0}\log^{1/2}k+(NT)^{-1/2}\log k\right\},
	\end{split}
\end{eqnarray}
with probability at least $1-k^{-2}(NT)^{-2}$. 

Next, we consider \eqref{eqn:eta6}. We note that \eqref{eqn:eta6} can be bounded by the sum of the following two terms,
\begin{eqnarray}\label{eqn:eta7}
	&&\left|\frac{2}{nT}\sum_{i\in \mathcal{I}_{\ell}}\sum_{t=0}^{T-1} \{Q_{i,t,a}-\Mean (Q_{i,t,a}|S_{i,t})\}\{\widehat{\tau}_a(S_{i,t})-\tau_a^*(S_{i,t})\}\right|,\\ \label{eqn:eta8}
	&&\left|\frac{2}{nT}\sum_{i\in \mathcal{I}_{\ell}}\sum_{t=0}^{T-1} \{\eta_{i,t,a}-\Mean (\eta_{i,t,a}|S_{i,t})\}\{\widehat{\tau}_a(S_{i,t})-\tau_a^*(S_{i,t})\}\right|,
\end{eqnarray}
where $Q_{i,t,a}$ denotes $\widehat{Q}^{(\ell)}(a,S_{i,t})+b^{-1}(a|S_{i,t})\mathbb{I}(A_{i,t}=a)\{R_{i,t}+\gamma \widehat{Q}^{(\ell)}(S_{i,t+1},\widehat{\pi}^{(\ell)}(S_{i,t+1}))-\widehat{Q}^{(\ell)}(A_{i,t},S_{i,t})\}$. 

Consider \eqref{eqn:eta7} first. Set $\mathcal{T}^*$ to be an $(nT)^{-1}$-net of $\mathcal{T}_{a,k}$ with respect to the distance measure $\|f_1-f_2\|_n=\sqrt{(nT)^{-1} \sum_{i\in \mathcal{I}_{\ell}}\sum_{t=0}^{T-1} |f_1(S_{i,t},a)-f_2(S_{i,t},a)|^2}$ for any $f_1$ and $f_2$. The number of elements in $\mathcal{T}^*$ is upper bounded by
\begin{eqnarray*}
	\mathcal{N}((nT)^{-1}, \mathcal{T}_{a,k}, \|\cdot\|_n)\le \mathcal{N}((nT)^{-1}, \mathcal{T}_{a,k}, \|\cdot\|_{\infty})\preceq \exp\left\{ O(1)W L\log W \log (nT)  \right\}\\\preceq \exp\left\{ O(1) (NT)^{\kappa_0d/p^*}\log^4 (NT) \right\}.
\end{eqnarray*}
It follows from Cauchy-Schwarz inequality that \eqref{eqn:eta7} is upper bounded by
\begin{eqnarray}\label{eqn:eta7.3}
	\sup_{\tau_a\in \mathcal{T}^*} \left|\frac{4}{nT}\sum_{i\in \mathcal{I}_{\ell}}\sum_{t=0}^{T-1} \{Q_{i,t,a}-\Mean (Q_{i,t,a}|S_{i,t})\}\{\tau_a(S_{i,t})-\tau_a^*(S_{i,t}) \}\right|+\frac{O(1)}{NT}.
\end{eqnarray}
For each $\tau_a\in \mathcal{T}^*$, $(nT)^{-1} \sum_{i\in \mathcal{I}_{\ell}}\sum_{t=0}^{T-1} \{Q_{i,t,a}-\Mean (Q_{i,t,a}|S_{i,t})\}\{\tau_a(S_{i,t})-\tau_a^*(S_{i,t})\}$ corresponds to a sum of martingale difference sequence. It follows from the Bernstein's inequality \citep[see e.g., Theorem A,][]{fan2015exponential} that with probability at least $1-\exp(u)$ that 
\begin{eqnarray*}
	\left|\frac{4}{nT}\sum_{i\in \mathcal{I}_{\ell}}\sum_{t=0}^{T-1} \{Q_{i,t,a}-\Mean (Q_{i,t,a}|S_{i,t})\}\{\tau_a(S_{i,t})-\tau_a^*(S_{i,t}) \}\right|\\\le O(1)(NT)^{-1/2}\max( k\sqrt{u}(NT)^{-\kappa_0}, (NT)^{-1/2}u ),
\end{eqnarray*}
for each $\tau_a\in \mathcal{T}^*$, under the event defined in $\mathcal{A}_k$. By Bonferroni's inequality, we obtain with probability at least $1-\exp\{O(1) (NT)^{\kappa_0d/p^*}\log^4 (NT)-u\}$ that the first term in \eqref{eqn:eta7.3} is upper bounded by
\begin{eqnarray*}
	%\sup_{f\in \mathcal{T}^*} \left|\frac{2}{nT}\sum_{i\in \mathcal{I}_{\ell}}\sum_{t=0}^{T-1} \{Q_{i,t,a}-\Mean (Q_{i,t,a}|S_{i,t})\}f(S_{i,t},a)\right|\le 
	O(1)(NT)^{-1/2}\max(k\sqrt{u}(NT)^{-\kappa_0},(NT)^{-1/2}u ).
\end{eqnarray*}
%Set $u$ to $\bar{C} (NT)^{\kappa_0d/p^*}\log^4 (NT)$ for some sufficiently large constant 
Combining this together with \eqref{eqn:eta7.3} yields that \eqref{eqn:eta7} is upper bounded by
\begin{eqnarray*}
	O(1)(NT)^{-1/2}\max(k\sqrt{u}(NT)^{-\kappa_0},(NT)^{-1}u )+O(1) (NT)^{-1}.
\end{eqnarray*}
Set $u$ to $\bar{C} (NT)^{\kappa_0d/p^*}\log^4 (NT)+2\log (kNT)$ for some sufficiently large constant $\bar{C}>0$, \eqref{eqn:eta7} is upper bounded by
\begin{eqnarray}\label{eqn:eta-2}
	\begin{split}
		\frac{O(1)}{\sqrt{NT}}\left\{k\sqrt{(NT)^{\kappa_0d/p^*-2\kappa_0} \log^4(NT)}+(NT)^{\kappa_0d/p^*-1/2}\log^4 (NT)\right.\\
		\left.+k(NT)^{-\kappa_0}\log^{1/2}k+(NT)^{-1/2}\log k\right\},
	\end{split}
\end{eqnarray}
with probability at least $1-k^{-2}(NT)^{-2}$.

Next let us consider \eqref{eqn:eta8}. Following the discussion below Lemma 4.1 of \cite{Dedecker2002},  we can construct a sequence of random variables $\{O_{i,t}^0\}_{i,t}$ such that $O_{i,t}^0$ has the same distribution function as $O_{i,t}$ and that
\begin{eqnarray}\label{eqn:zeta3ell4}
	\eta_{i,t,a}=\eta_{i,t,a}^0,
\end{eqnarray}
with probability at least $1-nT\beta(q)/q$, where $\eta_{i,t,a}^0$ corresponds to a version of $\eta_{i,t,a}$ with  $\{O_{i,t}\}_{0\le t\le T, i\in \mathcal{I}_{\ell}}$ replaced by $\{O_{i,t}^0\}_{0\le t\le T, i'\in \mathcal{I}_{\ell}}$. In addition, the sequences $\{U_{i,2t}^0\}_{i,t}$ and $\{U_{j,2t+1}^0\}_{i,t}$ are i.i.d. where $U_{i,t}^0=(O_{i,t(q+1)}^0,O_{i,t(q+1)+1}^0,\cdots,O_{i,t(q+1)+q}^0)$. 

%Using similar arguments in the proof of Theorem \ref{thm:1}, we can show
%\begin{eqnarray}\label{eqn:zeta3ell5}
%\begin{split}
%	&\Mean [\eta_{i,t,a,1}^{0*}-\eta_{i,t,a,1}^0|S_{i,t},\{O_{j,t}\}_{j\in \mathcal{I}_{\ell}^c,0\le t<T} ]\\
%	=&-\int_{s} \{V^{\widehat{\pi}^{(\ell)}}(s)-\widehat{V}^{(\ell)}(s)\}\mathcal{P}(ds;S_{i,t},a)+\epsilon_{i,t,1}, 
%\end{split}	
%\end{eqnarray}
%where $\epsilon_{i,t,1}$ satisfies $\max_{i,t}\Mean (|\epsilon_{i,t,1}||S_{i,t})=o(N^{-1/2} T^{-1/2})$ under the condition that $c_0>1/4$. 
By setting $q$ to be proportional to $\log (NT)$, the probability $1-nT\beta (q)/q$ can be lower bounded by $1-(NT)^{-1}$. %for some sufficiently large constant $C>0$. 
Without loss of generality, suppose $(n/\mathbb{L})T=2mq$ for some integer $m>0$. By definition, $I^0=\Mean (\eta_{i,t,a}^{0}-\eta_{i,t,a}^0|S_{i,t}^0)$ can be decomposed into three variables $I_{1}^0$, $I_{2}^0$ and $I_3^0$, where each of the first two corresponds to a sum of $m$ i.i.d. random variables with zero mean given $S_{i,t}$ and $\{O_{j,t}\}_{j\in \mathcal{I}_{\ell}^c,0\le t<T}$, and $|I_3^0|$ is a remainder term upper bounded by $O(1) (nT)^{-1} \log (nT)$. In addition, using similar arguments in Step 3 of the proof of Theorem \ref{thm:2}, we can show with probability at least $1-(nT)^{-C}$ that the bias 
\begin{eqnarray*}
	\max_{i,t} |\Mean (\eta_{i,t,a}|S_{i,t})-\Mean (\eta_{i,t,a}^0|S_{i,t}^0)|
\end{eqnarray*}
can be upper bounded by $O(1)(nT)^{-1}$ as well. Consequently, we obtain with probability $1-(NT)^{-1}$ that $	\max_{i,t} |\Mean (\eta_{i,t,a}|S_{i,t})-\Mean (\eta_{i,t,a}^0|S_{i,t}^0)|\le O(1) (nT)^{-1/2} \sqrt{\log (nT)}$.
Let $\mathcal{A}_0$ denote this event. On the set $\mathcal{A}_0$, using the Cauchy-Schwarz inequality,  \eqref{eqn:eta8} can be upper bounded by
\begin{eqnarray*}
	\frac{O(1)}{NT\varepsilon }\log (NT)+\varepsilon \left|\frac{2}{nT}\sum_{i\in \mathcal{I}_{\ell}}\sum_{t=0}^{T-1} |\widehat{\tau}_a(S_{i,t})-\tau_a^*(S_{i,t})|^2 \right|,
\end{eqnarray*}
for any sufficiently small $\varepsilon>0$. 

%Without loss of generality, suppose $\kappa<1/2$. 
To summarize, on the set $\mathcal{A}_0\cap \mathcal{A}^*\cap \mathcal{A}_k$, the stochastic error term \eqref{eqn:-1} can be upper bounded by
\begin{eqnarray*}
	&&\frac{O(1)(NT)^{-2\kappa}}{\varepsilon}+2\varepsilon\left|\frac{2}{nT}\sum_{i\in \mathcal{I}_{\ell}}\sum_{t=0}^{T-1} |\widehat{\tau}_a(S_{i,t})-\tau_a^*(S_{i,t})|^2 \right|\\
	&+& \frac{O(1)}{\sqrt{NT}}\left\{k\sqrt{(NT)^{\kappa_0d/p^*-2\kappa_0} \log^4(NT)}+(NT)^{\kappa_0d/p^*-1/2}\log^4 (NT)\right.\\
	&+&\left. k(NT)^{-\kappa_0}\log^{1/2}k+(NT)^{-1/2}\log k\right\},
\end{eqnarray*}
with probability at least $1-O(N^{-2}T^{-2}k^{-2})$. Similarly, we can show \eqref{eqn:var} is upper bounded by \begin{eqnarray}\label{eqn:varbound}
	\begin{split}
		&\frac{O(1)(NT)^{-2\kappa}}{\varepsilon}+2\varepsilon\left|\frac{2}{NT}\sum_{i=1}^N\sum_{t=0}^{T-1} |\widehat{\tau}_a(S_{i,t})-\tau_a^*(S_{i,t})|^2 \right|\\
		+& \frac{O(1)}{\sqrt{NT}}\left\{k\sqrt{(NT)^{\kappa_0d/p^*-2\kappa_0} \log^4(NT)}+(NT)^{\kappa_0d/p^*-1/2}\log^4 (NT)\right.\\
		+&\left. k(NT)^{-\kappa_0}\log^{1/2}k+(NT)^{-1/2}\log k\right\},
	\end{split}
\end{eqnarray}
with probability at least $1-O(N^{-2}T^{-2}k^{-2})$. 

By definition, we have
\begin{eqnarray*}
	\frac{1}{NT}\sum_{i=1}^N \sum_{t=0}^{T-1}\{\widetilde{\tau}_{i,t,a}-\tau_a^*(S_{i,t})\}^2\ge 	\frac{1}{NT}\sum_{i=1}^N \sum_{t=0}^{T-1}\{\widetilde{\tau}_{i,t,a}-\widehat{\tau}_a(S_{i,t})\}^2.
\end{eqnarray*}
Decomposing $\widetilde{\tau}_{i,t,a}-\widehat{\tau}_a(S_{i,t})$ into the sum of $\widetilde{\tau}_{i,t,a}-\tau_a^*(S_{i,t})$ and $\tau_a^*(S_{i,t})-\widehat{\tau}_a(S_{i,t})$, the stochastic error term shall be smaller than the squared bias term
\begin{eqnarray}\label{eqn:-4}
	\frac{1}{NT}\sum_{i=1}^N \sum_{t=0}^{T-1}\{\tau_a^*(S_{i,t})-\widehat{\tau}_a(S_{i,t})\}^2.
\end{eqnarray}
Set $\varepsilon$ in \eqref{eqn:varbound} to $1/8$. 
On the set $\mathcal{A}_0\cap \mathcal{A}^* \cap \mathcal{A}_k$, we obtain with probability at least $1-O\{(NTk)^{-1}\}$ that
\begin{eqnarray}\label{eqn:-3}
	\begin{split}
		&\frac{1}{8NT}\sum_{i=1}^N \sum_{t=0}^{T-1}\{\tau_a^*(S_{i,t})-\widehat{\tau}_a(S_{i,t})\}^2-O(1) (NT)^{-2\kappa}\\ \le& 
		\frac{O(1)}{\sqrt{NT}}\left\{k\sqrt{(NT)^{\kappa_0d/p^*-2\kappa_0} \log^4(NT)}+(NT)^{\kappa_0d/p^*-1/2}\log^4 (NT)\right.\\
		\le &\left.+k(NT)^{-\kappa_0}\log^{1/2}k+(NT)^{-1/2}\log k\right\}.
	\end{split}	
\end{eqnarray}
Using similar arguments in proving \eqref{eqn:varbound}, we can show the left-hand-side (LHS) is larger than
\begin{eqnarray*}
	&&\frac{1}{8}\Mean |\tau_a^*(S_0)-\widehat{\tau}_a(S_0)|^2-O(1)(NT)^{-2\kappa}\\
	&-&\frac{O(1)}{\sqrt{NT}}\left\{k\sqrt{(NT)^{\kappa_0d/p^*-2\kappa_0} \log^4(NT)}+(NT)^{\kappa_0d/p^*-1/2}\log^4 (NT)\right.\\
	&+&\left.k(NT)^{-\kappa_0}\log^{1/2}k+(NT)^{-1/2}\log k\right\},
\end{eqnarray*}
with probability at least $1-O\{(NTk)^{-2}\}$, on the set $\mathcal{A}_0\cap \mathcal{A}^* \cap \mathcal{A}_k$. In particular, note that the first term is at least $k^2(NT)^{-2\kappa_0}/8$ under $\mathcal{A}_k$. By setting $\kappa_0=\min\{\kappa,p^*/(2p^*+d)\}$, for sufficiently large $k_0$, the above expression is strictly larger than the right-hand-side of \eqref{eqn:-3}. This violates the result in \eqref{eqn:-4}. Consequently, on the set $\mathcal{A}_0$, the event $\mathcal{A}_k^c$ holds with probability at least $1-O\{(NTk)^{-2}\}$. By Bonferroni's inequality, on the set $\mathcal{A}_0\cap \mathcal{A}^*$, the event $\cap_{k\ge k_0}\mathcal{A}_k^c$ holds with probability at least
\begin{eqnarray*}
	1-\sum_{k\ge k_0}O\{(NTk)^{-2}\}=1-O(N^{-2}T^{-2}). 
\end{eqnarray*}
Since both $\mathcal{A}_0$ and $\mathcal{A}^*$ hold with probability at least $1-O(N^{-1}T^{-1})$, \eqref{eqn:upperboundprob} occurs with probability at most $O(N^{-1} T^{-1})$. The proof is thus completed.

%\appendix
\section{Some implementation details}\label{sec:weight}
%\subsection{Fitted-Q evaluation}
We first present the fitted-Q evaluation algorithm below in Algorithm \ref{FQE}.  
\begin{algorithm}[!t]
	\caption{Fitted-Q evaluation}\label{FQE}
	\begin{algorithmic}
		\STATE \textbf{Input:} a testing dataset $\{S_{j,t},A_{j,t},R_{j,t},S_{j,t+1}\}_{j,t}$, an estimated optimal policy $\pi$, a class of function approximators $\mathcal{F}$ and a discounted factor $\gamma$;
		\STATE Randomly pick a $Q_0 \in \mathcal{F}$\;
		\STATE \textbf{For} {$k = 1, \dots, K$} 
		
		\STATE Update target values $Z_{j,t} = R_{j,t} + \gamma Q_{k-1}(S_{j,t+1}, \pi(S_{j,t+1}))$ for all $(j,t)$;
		
		\STATE Solve a regression problem to update the $Q$-function:\\
		\quad \quad    $Q_k = \argmin_{Q \in \mathcal{F}} \frac{1}{n} \sum_{i=1}^n \{Q(S_{j,t}, A_{j,t}) - Z_{j,t}\}^2$
		\STATE \textbf{Output:} The estimated value $\widehat{V}(\cdot)=Q_K(\cdot,\pi(\cdot))$
	\end{algorithmic}	
\end{algorithm}

%\subsection{Estimation of the density ratio}
We next discuss estimation of the density ratio. Under the stationarity assumption in (A5), we have
\begin{eqnarray*}
	\omega^{\pi}(a',s'|a,s)=\frac{\pi(a'|s')}{b(a'|s')}\underbrace{\frac{(1-\gamma)\sum_{t\ge 1}\gamma^{t-1}p_t^{\pi}(s'|a,s)}{p_{\infty}(s)}}_{w^{\pi}(s'|a,s)},
\end{eqnarray*}
where $p_t^{\pi}(s'|a,s)$ denotes the conditional probability density function of $S_t$ given the initial state action pair $(a,s)$ assuming the system follows $\pi$ at time $1,2,\cdots,t-1$. This motivates us to compute $\widehat{\omega}^{(\ell)}$ by 
\begin{eqnarray*}
	\frac{\pi(a'|s')}{b(a'|s')}\widehat{w}^{(\ell)}(s'|a,s),
\end{eqnarray*}
where $\widehat{w}^{(\ell)}$ denotes some consistent estimator for $w^{\widehat{\pi}^{(\ell)}}$ based on the data subset in $\mathcal{I}_{\ell}^c$. Our procedure for computing $\widehat{w}^{(\ell)}$ is motivated by the following lemma. 
\begin{lemma}\label{lemma:1}
	For any two pairs $(i,t)$ and $(i',t')$ such that $(S_{i,t},A_{i,t},S_{i,t+1})$ and $(S_{i',t'},A_{i',t'},S_{i',t'+1})$ are independent, we have for any policy $\pi$ and any function $f$ that \vspace{-0.1cm}
	%Let $\widetilde{O}_{i,t}=(\widetilde{S}_{i,t},\widetilde{A}_{i,t},\widetilde{R}_{i,t},\widetilde{S}_{i,t+1})$ be an independent copy of $O_{i,t}$. Then for any function $f$, we have \vspace{-0.1cm}
	\begin{eqnarray}\label{eqn:omega}
		\begin{split}
			&\gamma \Mean\{ w^{\pi}(S_{i',t'}|A_{i,t},S_{i,t})\frac{\pi(A_{i',t'}|S_{i',t'})}{b(A_{i',t'}|S_{i',t'})}f(S_{i',t'+1},A_{i,t},S_{i,t})\}\\
			-&\Mean\{ w^{\pi}(S_{i',t'+1}|A_{i,t},S_{i,t})f(S_{i',t'+1},A_{i,t},S_{i,t})\}
			=-(1-\gamma) \Mean\{ f(S_{i,t+1},A_{i,t},S_{i,t})\}.
		\end{split}	
	\end{eqnarray}
\end{lemma}
For each function $f$, an estimating equation for $w^{\pi}$ can be constructed based on  \eqref{eqn:omega}. Similar to the proposal in \cite{liu2018}, we treat $f$ as a discriminator to construct a mini-max loss
function. %Then the density ratio is estimated by optimizing this loss function. 

Specifically, denote by $\Delta(w^{\pi};i,t, i',t')$ to be the left-hand-side (LHS) of \eqref{eqn:omega}. 
Consider the following optimization problem
\begin{eqnarray}\label{optimize}\\\nonumber
	\argmin_{w^{\widehat{\pi}^{(\ell)}}\in \mathcal{W}} \sup_{f\in \mathcal{F}} \left|\sum_{\substack{(i,i')\in \mathcal{I}_{\ell}^c\\ (i,t)\neq (i',t')}} \Delta(w^{\widehat{\pi}^{(\ell)}};i,t,i',t')f(S_{i',t'+1},A_{i,t},S_{i,t})+(1-\gamma) f(S_{i,t+1},A_{i,t},S_{i,t})\right|^2
\end{eqnarray}
In our implementation, we set $\mathcal{F}$ to a unit ball of a reproducing kernel Hilbert space (RFHS), i.e., 
\begin{eqnarray*}
	\mathcal{F}=\{f\in \mathcal{H}:\|f\|_{\mathcal{H}}=1\},
\end{eqnarray*}
where %$\mathcal{H}$ corresponds to an RFHS such that
\begin{eqnarray*}
	\mathcal{H}=\left\{f(\cdot)=\sum_{i,i'\in \mathcal{I}_{\ell}^c, (i,t)\neq (i',t')} b_{i,t,i',t'} \kappa(S_{i',t'},A_{i,t},S_{i,t};\bullet): b_{i,t,i',t'}\in \mathbb{R} \right\},
\end{eqnarray*}
for some positive definite kernel $\kappa(\cdot;\cdot)$. Similar to Theorem 2 of \cite{liu2018}, the optimization problem in \eqref{optimize} is then reduced to
\begin{eqnarray*}
	\argmin_{w^{\widehat{\pi}^{(\ell)}}\in \mathcal{W}} \frac{1}{\displaystyle{\sum_{i\in \mathcal{I}_{\ell}^c}T_i\choose 2}^2} \sum_{\substack{i_1,i_1'\in \mathcal{I}_{\ell}^c\\ (i_1,t_1)\neq (i_1',t_1')}}\sum_{\substack{i_2,i_2'\in \mathcal{I}_{\ell}^c\\ (i_2,t_2)\neq (i_2',t_2')}} D(w^{\widehat{\pi}^{(\ell)}};i_1,t_1,i_1',t_1',i_2,t_2,i_2',t_2'),
\end{eqnarray*}
where $D(w^{\widehat{\pi}^{(\ell)}};i_1,t_1,i_1',t_1',i_2,t_2,i_2',t_2')$ is given by
\begin{eqnarray*} \Delta(w^{\widehat{\pi}^{(\ell)}};i_1,t_1,i_1',t_1')\Delta(w^{\widehat{\pi}^{(\ell)}};i_2,t_2,i_2',t_2') \kappa(S_{i_1',t_1'+1},S_{i_1,t_1},A_{i_1,t_1};S_{i_2',t_2'+1},S_{i_2,t_2},A_{i_2,t_2})\\
	+(1-\gamma) \Delta(w^{\widehat{\pi}^{(\ell)}};i_1,t_1,i_1',t_1') \kappa(S_{i_1',t_1'+1},S_{i_1,t_1},A_{i_1,t_1};S_{i_2,t_2+1},S_{i_2,t_2},A_{i_2,t_2})\\
	+(1-\gamma) \Delta(w^{\widehat{\pi}^{(\ell)}};i_2,t_2,i_2',t_2') \kappa(S_{i_1,t_1+1},S_{i_1,t_1},A_{i_1,t_1};S_{i_2',t_2'+1},S_{i_2,t_2},A_{i_2,t_2})\\
	+(1-\gamma)^2 \kappa(S_{i_1,t_1+1},S_{i_1,t_1},A_{i_1,t_1};S_{i_2,t_2+1},S_{i_2,t_2},A_{i_2,t_2})
\end{eqnarray*}
We set $\mathcal{W}$ to a class of neural networks. %One could use different parameters to factorize different $\omega_i$ such that each $\widehat{\omega}_i$ is computed separately. Alternatively, one could allow different $\omega_i$ to share some common parameters. We detail our
The detailed estimating procedure is given in Algorithm \ref{alg1}. 

\begin{algorithm}[t!]
	\caption{Estimation of the density ratio.}
	\label{alg1}
	%\vspace*{-15pt}
	\begin{algorithmic}
		\item
		\begin{description}
			\item[\textbf{Input}:] The data $\{(S_{i,t},A_{i,t},S_{i,t+1}):i\in \mathcal{I}_{\ell}^c,0\le t<T_i\}$. An estimated optimal policy $\widehat{\pi}^{(\ell)}$. 
			
			\item[\textbf{Initial}:] Initial the density ratio $w=w_{\beta}$ to be a neural network parameterized by $\beta$.
			
			\item[\textbf{for}] iteration $=1,2,\cdots$ \textbf{do}
			\begin{enumerate}
				\item[a] Randomly sample batches $\mathcal{M}$, $\mathcal{M}^*$ from $\{(i,t): i\in \mathcal{I}_{\ell}^c, 0\le t<T_i \}$.
				
				\item[b] {\textbf{Update}} the parameter $\theta$ by $$\beta\leftarrow \beta-\epsilon {|\mathcal{M}|\choose 2}^{-2}\sum_{\substack{(i_1,t_1),(i_1',t_1')\in \mathcal{M}\\ (i_1,t_1)\neq (i_1',t_1') }}\sum_{\substack{(i_2,t_2),(i_2',t_2')\in \mathcal{M}\\ (i_2,t_2)\neq (i_2',t_2') }} \nabla_{\beta} D(w_{\beta}/z_{w_{\beta}};i_1,t_1,i_1',t_1',i_2,t_2,i_2',t_2'),$$ 
				where $z_{w_{\beta}}$ is a normalization constant $$z_{w_{\beta}}(\cdot;S_{i,t},A_{i,t})=\frac{1}{|\mathcal{M}^*|} \sum_{(i',t') \in \mathcal{M}^* } w_{\beta}(S_{i',t'};A_{i,t},S_{i,t}).$$ 
			\end{enumerate}
			\item[\textbf{Output}] $w_{\beta}$. 
		\end{description}
	\end{algorithmic}
\end{algorithm}

Finally, we present the detailed definition of $A_t$ and $R_t$ in the real data application. Specifically, 
\begin{align*}
	A_t  = \begin{cases}
		0, & \text{In}_t = 0; \\
		m, & 4m-4 < \text{In}_t \le 4m \, (m=1, 2, 3); \\
		4, & \text{In}_t > 12,
	\end{cases}
\end{align*}
and 
\begin{align*}
	R_t = \begin{cases}
		-\frac{1}{30}(80-G_{t+1})^2, & G_{t+1} < 80; \\
		0, & 80 \le G_{t+1} \le 140; \\
		-\frac{1}{30}(G_{t+1} - 140)^{1.35}, & 140 \le G_{t+1}.
	\end{cases}
\end{align*}

\end{document}